%% file: main.tex
\documentclass[10pt,twocolumn,letterpaper]{article}

\usepackage[pagenumbers]{cvpr} 

\input{preamble}
\usepackage{placeins}
\definecolor{cvprblue}{rgb}{0.21,0.49,0.74}
\usepackage[pagebackref,breaklinks,colorlinks,allcolors=cvprblue]{hyperref}

\def\paperID{*****} 
\def\confName{CVPR}
\def\confYear{2026}

\title{CineRecon: Benchmarking Frontier Models\\for Agentic Cinematic Reconstruction}

\author{
Jialei Wang$^{1}$\quad Zehan Wang$^{2,\dagger}$\quad XiaoYang Wu$^{3}$\quad Ziang Zhang$^{2}$\quad Siyu Chen$^{2}$\\
Jiaer Huang$^{1}$\quad Yujia Zhang$^{1}$\quad Tianyu Pang\quad Hengshuang Zhao$^{1,*}$\\
{$^{1}$The University of Hong Kong\quad $^{2}$Zhejiang University\quad $^{3}$New York University}\\
{\href{https://cinerecon.github.io/}{\texttt{cinerecon.github.io}}}
}

\begin{document}
\maketitle
\begingroup
\renewcommand{\thefootnote}{}
\footnotetext[0]{$^{\dagger}$Project Lead\quad $^{*}$Corresponding Author}
\endgroup
\input{fig/teaser}
\raggedbottom

\begin{abstract}
Reconstructing reference videos as editable dynamic 3D scenes is an important path toward controllable video generation for filmmaking: these scenes provide modifiable, renderable visual proxies that enable explicit control over scene layout, subject actions, and camera behavior.
To evaluate agents' ability to reconstruct editable dynamic 3D scenes from dynamic reference videos, we introduce \textbf{CineRecon}, a benchmark comprising 50 reference videos and 360 annotated evaluation clips.
The reference videos span film, real-world footage, and games.
Given a complete video, an agent uses tools to construct, animate, render, and refine its reconstruction.
We evaluate this task under three representation settings: \textbf{Proxy}, \textbf{Simplified}, and \textbf{Detailed}.
These settings increase geometric and appearance requirements while retaining the same requirements for spatial relations, camera behavior, and subject dynamics.
This design tests whether agents can reproduce the events in a video when detailed appearance is unnecessary, and how performance changes as representation requirements increase.
For evaluation, we construct checklists that associate observable properties and events with specific entities and times.
A vision-language judge compares timestamp-aligned reference and rendered frames across 11 categories, grouped into camera, static, and dynamic consistency dimensions.
We evaluate 12 models using their corresponding agent harnesses and find failures across all three dimensions.
Subject motions and interactions remain difficult to reproduce even under the Proxy setting.
\end{abstract}

\section{Introduction}
\label{sec:introduction}

Precisely expressing and refining creative intent is a central requirement for bringing video generation into filmmaking~\cite{wang2024motionctrl,he2024cameractrl}.
Reconstructing reference videos as editable dynamic 3D scenes provides an important path toward this goal.
These scenes let creators explicitly adjust spatial relations, action timing, and camera trajectories by editing layout, subject actions, and camera behavior. Their rendered outputs serve as visual proxies to guide video generation~\cite{hu2024scenecraft,liu2025shape,fang2026spatialcrafter}.
Building such proxies requires faithfully reproducing the dynamic processes in the reference video.
For example, a camera follows a person approaching another person, and the two then shake hands.
The agent must coordinate the person's movement, the handshake, and the camera motion so that the interaction occurs at the correct time and location.
This raises a concrete question: \emph{can agents autonomously construct editable 3D scenes from complete videos to faithfully reenact their content and events?}

Recent work has expanded the evaluation of graphics agents from scene editing and static reconstruction~\cite{gu2025blendergym,yin2026vision,liu20263dharnessbench} to video reconstruction.
BVB studies real indoor videos~\cite{tang2026bvb}, while 4DCodeBench examines dynamic phenomena such as deformation, fluid flow, and fracture~\cite{shen20264dcodebench}.
Cinematic reconstruction requires jointly reproducing scene content, subject actions and interactions, camera motion, and shot transitions.
We evaluate this joint capability using complete reference videos, without supplied scene assets or camera parameters.

We introduce \textbf{CineRecon}, a benchmark of 50 reference videos from film, real-world footage, and games.
An agent receives the complete video and generic task instructions, then autonomously constructs an editable Blender scene and renders its animation.
The agent can inspect rendered feedback and revise the scene.
Success requires faithfully reproducing the reference video's content and events at corresponding timestamps.

We evaluate reconstruction along three dimensions. \textbf{Camera Consistency} covers camera behavior and shot transitions. \textbf{Static Consistency} covers object presence, attributes, and spatial relations. \textbf{Dynamic Consistency} covers subject motion and action events, including interactions, deformation, and state changes, with attention to their timing.
For each reference video, we construct checklists that associate observable properties and events with specific entities and times.
A vision-language judge compares timestamp-aligned reference and rendered frames to assess these requirements.

To examine how geometric and appearance requirements affect reconstruction performance, we define three settings for the same reference videos: \textbf{Proxy}, \textbf{Simplified}, and \textbf{Detailed}.
They require geometric stand-ins, recognizable object structures, and reference-specific visible details, respectively.
Geometric and appearance-detail requirements within Static Consistency vary across settings, while Camera and Dynamic Consistency use the same check items and spatial-relation requirements remain unchanged.
This design allows us to compare reconstruction performance across representation settings and examine whether errors in reproducing dynamic events persist when appearance requirements are simplified.

\begin{figure*}[t]
\centering
\includegraphics[width=\textwidth]{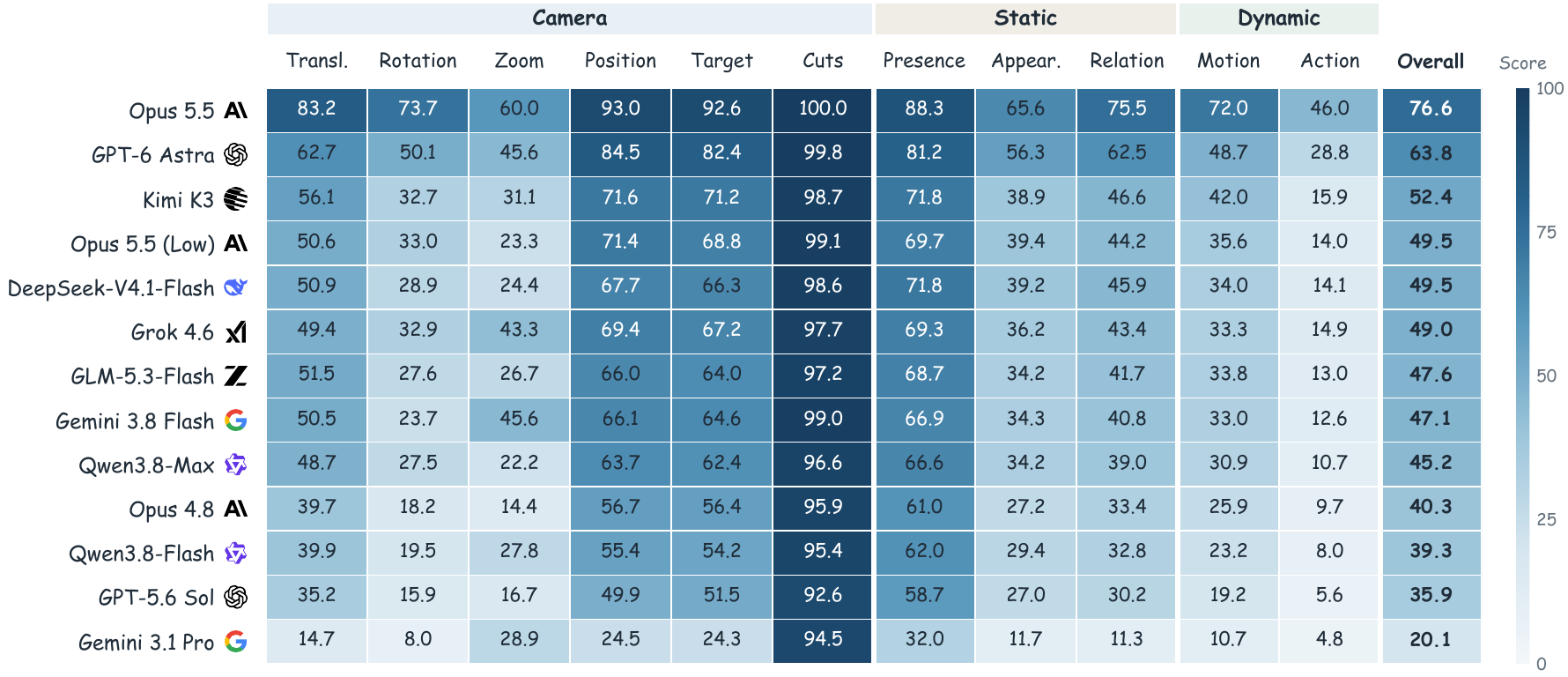}
\caption{\textbf{Reconstruction performance.} Scores for 13 configurations of 12 models across 11 check categories and Overall, averaged equally across Proxy, Simplified, and Detailed. Appearance averages Simplified and Detailed. Overall weights camera, static, and dynamic consistency equally. Low denotes low reasoning effort.}
\label{fig:category_scores}
\end{figure*}

\begin{figure}[tbp]
\centering
\includegraphics[width=\linewidth]{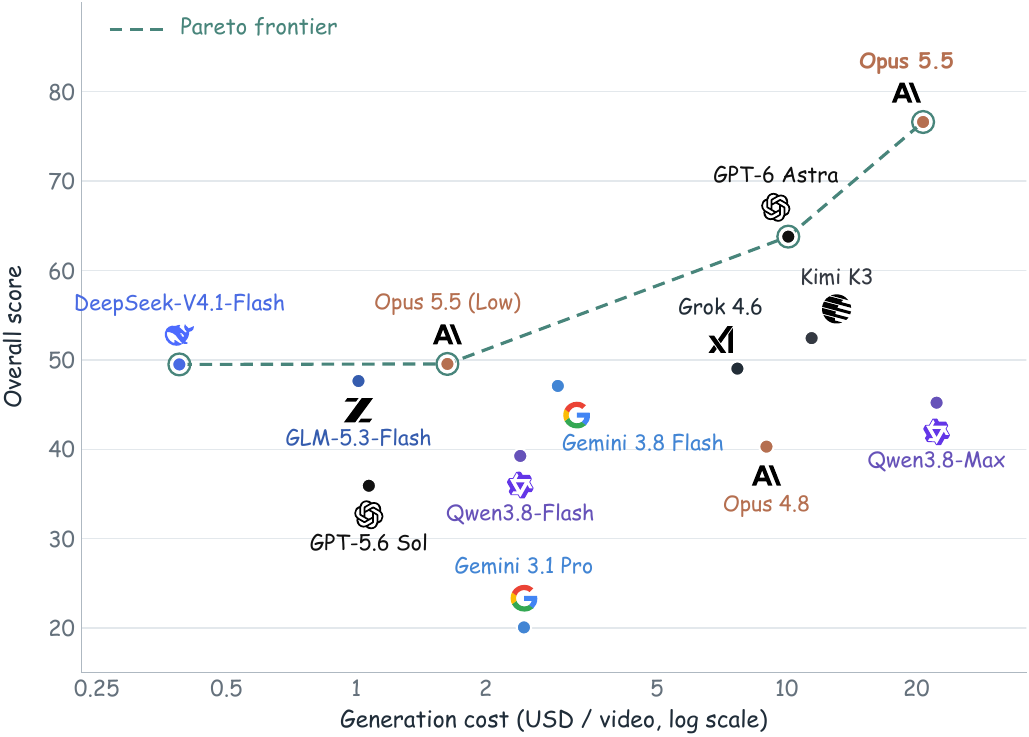}
\caption{\textbf{Generation cost and reconstruction performance.} Overall score versus mean generation cost per video (USD, logarithmic axis). Scores weight the three consistency dimensions and three representation settings equally. The dashed line connects the observed Pareto frontier. Low denotes low reasoning effort.}
\label{fig:score_cost}
\end{figure}

We systematically evaluate existing multimodal coding models. Figures~\ref{fig:category_scores} and~\ref{fig:score_cost} summarize reconstruction performance and generation cost.
The results identify spatial relations, specific action events, and camera changes as shared weaknesses across models.
Cross-setting comparisons show that most models achieve higher camera and dynamic consistency scores when appearance requirements are simplified, but still struggle to reproduce action timing, contact relations, and coordinated motion between subjects accurately.

In summary, our contributions are threefold:
(1) We introduce CineRecon, a benchmark comprising 50 reference videos and 360 annotated evaluation clips for evaluating agents that reconstruct complete dynamic videos as editable Blender scenes.
(2) We define three representation settings and evaluate camera, static, and dynamic consistency using evaluation clips and checklists that have undergone multiple rounds of human review and refinement.
(3) We analyze how 12 multimodal coding models perform with their corresponding agent harnesses and identify shared deficits in spatial relations, specific action events, and camera changes. Cross-setting comparisons further show that simplifying appearance requirements improves performance for most models, while substantial errors in action timing, contact relations, and coordinated subject motion persist under Proxy.

\begin{figure*}[!t]
  \centering
  \includegraphics[width=\linewidth]{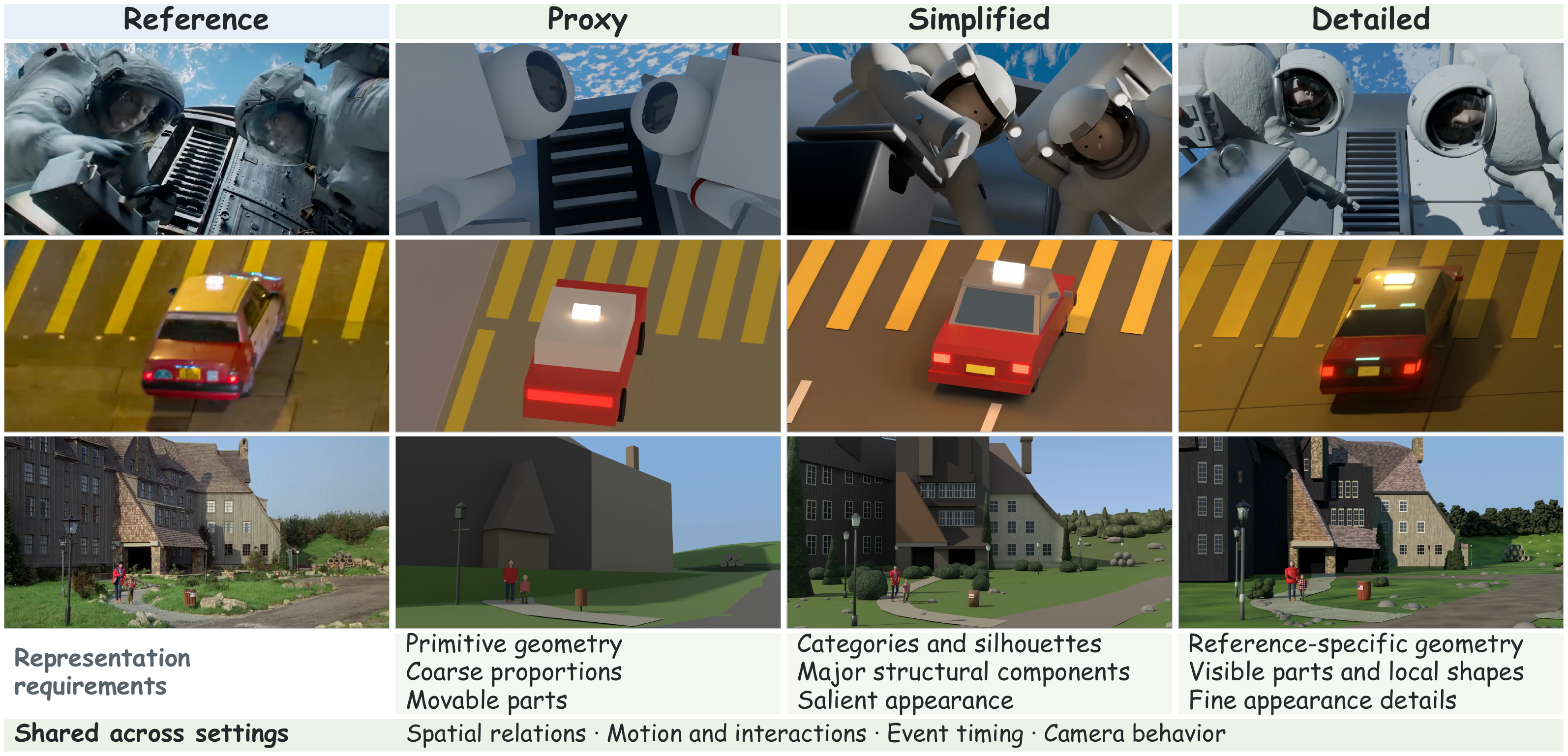}
  \caption{\textbf{Representation settings.} Reference frames and Opus 5.5 reconstructions under Proxy, Simplified, and Detailed. Geometric and appearance requirements increase across the three settings, while requirements for spatial relations, camera behavior, motion, interactions, and event timing remain unchanged.}
  \label{fig:representation_settings}
\end{figure*}

\section{CineRecon}

\subsection{Task Formulation}
\label{sec:task_formulation}

Given a complete reference video $V$ and a representation setting $r \in \{\mathrm{Proxy},\mathrm{Simplified},\mathrm{Detailed}\}$, an agent $\mathcal{A}$ constructs an editable Blender scene $B$ and renders a video $\hat{V}$:
\begin{equation}
  B = \mathcal{A}(V,r), \qquad \hat{V} = \mathcal{R}(B),
  \label{eq:reconstruction_task}
\end{equation}
where $\mathcal{R}$ renders the scene along its animation timeline.
The scene specifies geometry, appearance, subject animation, and camera configuration.
The objective is to reproduce the observable reference content at corresponding timestamps while satisfying setting $r$.
The output must match the reference resolution, frame rate, frame count, and duration.
For multi-shot videos, the camera reconstruction includes shot transitions.

Figure~\ref{fig:benchmark_overview} illustrates this reconstruction workflow, with aligned reference and rendered frames and the resulting editable scene.
The agent inspects the reference, constructs and animates the scene, and uses rendered feedback to revise its reconstruction.
It receives generic task instructions, but no video-specific annotations, reference camera parameters, or evaluation checklists.
The main protocol prohibits external assets, asset-generation services, and the use of reference imagery as backgrounds or projected surfaces in place of 3D modeling.
We assess reconstructed scenes by comparing their rendered videos with the visible content of the reference videos.

\begin{figure*}[!t]
\centering
\includegraphics[width=\linewidth]{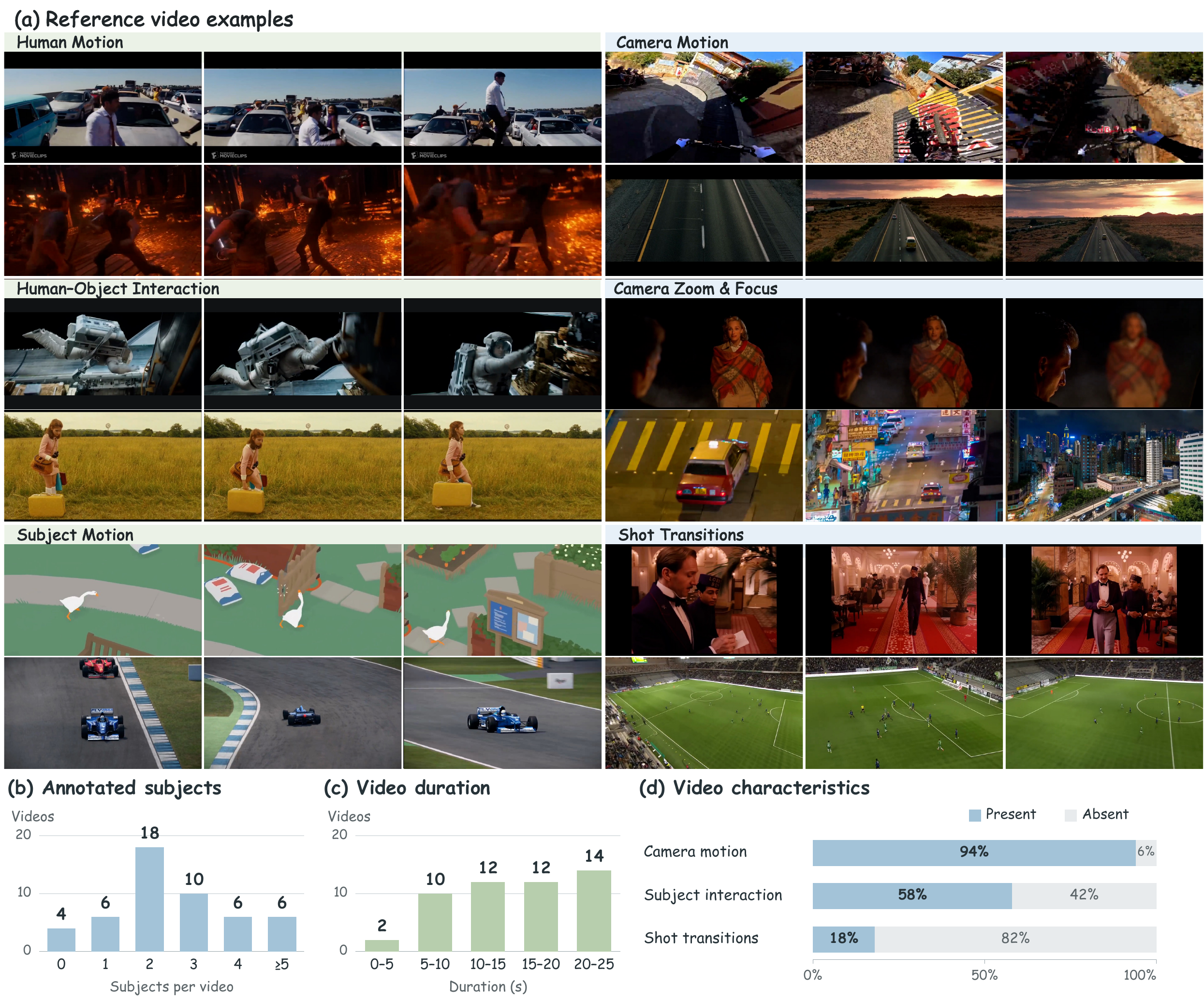}
\caption{\textbf{Reference videos and dataset characteristics.} (a) Reference sequences illustrating subject actions, motion and interactions, camera motion, zoom and focus changes, and shot transitions. (b) Annotated subjects per video: individually numbered characters, animals, and other main objects, counted once across clips within each video. (c) Reference video duration. (d) Video-level coverage of camera motion, subject interaction, and shot transitions, aggregated from the annotations. Statistics cover all 50 reference videos. The characteristics in (d) may co-occur.}
\label{fig:dataset_distribution}
\end{figure*}

\subsection{Representation Settings}
\label{sec:representation_settings}

The three settings increase geometric and appearance requirements while preserving requirements for spatial relations, camera behavior, motion, interactions, and event timing (Figure~\ref{fig:representation_settings}).
Each setting specifies the geometric and appearance features that the output should preserve.

\paragraph{Proxy.}
Subjects can be represented by a small number of geometric primitives or simple combinations.
The construction must preserve subject counts, coarse proportions, and the movable parts needed to express the observed actions.
Category-specific shapes, colors, clothing, materials, textures, and decorations are not required.
For articulated motion, the proxy must contain suitable moving parts when a rigid object would be unable to reproduce the action.

\paragraph{Simplified.}
Subjects must have recognizable categories, overall silhouettes, and major structural components.
Proportions and salient appearance features should broadly match the reference.
Small textures, ornaments, and other fine details may be simplified.

\paragraph{Detailed.}
Subjects must reproduce the visible structure and appearance of the specific reference instances, including distinctive local shapes, clearly visible parts, and fine appearance features.

\subsection{Data Collection and Annotation}
\label{sec:data_collection}

CineRecon contains 50 reference videos, including film footage, sports and first-person recordings, aerial footage, gameplay, and racing.
The collection includes both captured and rendered content, with subject motions and interactions, camera movement, changing viewpoints, and shot transitions.
Each complete video is a reconstruction input. Temporal annotations are used only for evaluation.
Figure~\ref{fig:dataset_distribution} illustrates the reference sequences and distributions of annotated subject counts and video duration.
Aggregated by reference video, the annotations record camera motion in 94\% of videos, subject interaction in 58\%, and shot transitions in 18\%.

We annotate 360 evaluation clips from the 50 reference videos and construct a structured checklist for each clip.
An evaluation clip spans a contiguous interval of at most two seconds and captures a salient scene state, camera change, subject motion, or interaction.
Clips may overlap and may cross shot boundaries.
All evaluation clips and checklist statements were initially annotated by Codex powered by GPT-6 Astra, followed by multiple rounds of human review and refinement.

Within each clip, each checklist statement associates an observable property or event with an identifiable entity and a specified timestamp or interval.
The annotations cover 11 categories across camera, static, and dynamic consistency.
Applicable statements are adapted to the representation setting.
Detailed dataset statistics are provided in Appendix~\ref{sec:dataset_statistics}.

\subsection{Evaluation Protocol}
\label{sec:evaluation_protocol}

We evaluate the rendered reconstruction against its reference video using the annotated checklists.
A vision-language judge compares reference and rendered frames at aligned timestamps and assesses the applicable statements under the prescribed representation setting.
The 11 checklist categories are organized into three consistency dimensions.

\textbf{Camera Consistency} covers six categories: \emph{translation}, the direction of camera displacement relative to its orientation at the start of the checked interval; \emph{rotation}, changes in pan, tilt, and roll; \emph{zoom}, visible zoom-in or zoom-out effects associated with changes in field of view; \emph{position}, the camera's location relative to scene landmarks at a specified frame; \emph{target}, the subject or region toward which the camera points at that frame; and \emph{shot transitions}, the number of shot transitions within the clip. Translation, rotation, and zoom are checked over intervals within a shot. Position and target describe the viewpoint at individual timestamps.

\textbf{Static Consistency} covers three categories at specified frames: \emph{object presence}, whether the required entities or scene regions are visible in the correct numbers; \emph{object appearance}, whether their shape, color, material, and visible details meet the representation requirements; and \emph{spatial relations}, whether relative positions, orientations, occlusions, and support or contact relationships are reproduced. Appearance checks apply to Simplified and Detailed, with detail requirements defined by the setting.

\textbf{Dynamic Consistency} covers two categories: \emph{subject motion}, the direction and path of overall displacement relative to a specified reference, including approaching, receding, and entering or leaving view; and \emph{action events}, events beyond whole-subject displacement within a specified time interval, such as arm raising, grasping, opening or closing, as well as interactions, deformation, and state changes.

For each dimension, we average check-item scores within each clip and then average across clips with equal weight. The overall score is the arithmetic mean of the three dimension scores. Since check-item counts differ substantially across dimensions, equal weighting prevents dimensions with more annotations from dominating the overall score.
The score measures agreement with the reference in the rendered evidence.
\section{Experiments}
\label{sec:experiments}

\subsection{Experimental Setup}
\label{sec:experimental_setup}

We evaluate 12 multimodal coding models: GPT-6 Astra~\cite{openai2026astra}, GPT-5.6 Sol~\cite{openai2026sol}, Claude Opus 5.5~\cite{anthropic2026opus55}, Claude Opus 4.8~\cite{anthropic2026opus48}, Gemini 3.1 Pro~\cite{google2026gemini31}, Gemini 3.8 Flash~\cite{google2026gemini38}, Qwen3.8-Max~\cite{alibaba2026qwenmax}, Qwen3.8-Flash~\cite{alibaba2026qwenflash}, Kimi K3~\cite{moonshot2026kimik3}, Grok 4.6~\cite{xai2026grok46}, DeepSeek-V4.1-Flash~\cite{deepseek2026v41}, and GLM-5.3-Flash~\cite{zai2026glm53}. Qwen3.8-Max and Qwen3.8-Flash use xhigh reasoning effort. The other ten models use high. We additionally include Opus 5.5 at low effort, giving 13 configurations in the quantitative comparisons. Model names without an effort label refer to the configurations listed above. Opus 5.5 (Low) denotes the low-effort configuration. Each configuration reconstructs the 50 reference videos under Proxy, Simplified, and Detailed. The Opus 5.5 ablation compares its high- and low-effort configurations with an additional, ablation-only No Refine variant (Sec.~\ref{sec:iteration_effort}).

Each model operates through its corresponding agent harness, with Blender 5.2~\cite{blender2026}, Blender MCP~\cite{ahuja2025blendermcp}, and file, shell, and media-inspection tools. GPT models use Codex CLI, Opus models use Claude Code, and both Gemini models use Antigravity CLI. Appendix~\ref{sec:model_configuration} lists the complete configuration. Each task provides the full reference video and a setting-specific reconstruction prompt. The agent selects observations, constructs and animates the scene, and inspects rendered feedback to guide further edits. It submits an editable Blender project and a video matching the reference resolution and timeline. Evaluation clips and checklists are withheld during reconstruction. The output limit is 32,768 tokens per model response. The agent chooses its refinement rounds. Prompts and execution details appear in Appendix~\ref{sec:appendix_configuration}.

\subsection{Automatic Evaluation}
\label{sec:automatic_evaluation_setup}

Multimodal evaluators assess instruction adherence in image synthesis~\cite{ku2024viescore} and predict multidimensional human ratings of generated videos~\cite{he2024videoscore}.
We use GPT-5.6 Sol~\cite{openai2026sol} with high reasoning effort as the VLM judge. Each request evaluates one reconstructed clip under one representation setting, using timestamp-aligned reference and rendered frames, the reference entity table, and the applicable checklist. Frames cover annotated observations, event intervals, and shot transitions. The judge rates each item as unsatisfied, partially satisfied, or fully satisfied, mapped to 0, 0.5, and 1. We aggregate scores over the 360 annotated clips following Sec.~\ref{sec:evaluation_protocol} and report camera, static, dynamic, and overall scores on a 0--100 scale. Appendix~\ref{sec:appendix_automatic_evaluation} specifies frame selection, judging instructions, and score aggregation.

\subsection{Human Evaluation}
\label{sec:human_evaluation_setup}

We assess alignment with human judgments through dimension-specific pairwise preferences. The study comprises 1,440 comparisons across 12 models, with 480 each for camera, static, and dynamic consistency. Model reasoning settings follow Table~\ref{tab:agent_configuration}. Each task presents a reference clip or frame and two anonymized reconstructions under the same representation setting. Raters select the result that better matches the reference on the specified dimension or indicate no clear difference. Human and automatic win rates use the same comparisons, count ties as half a win, and weight the three settings equally within each dimension. The overall win rate averages the three dimensions equally. Sampling, annotation, and preference aggregation are detailed in Appendix~\ref{sec:appendix_human_evaluation}.

\begin{figure}[!htb]
\centering
\includegraphics[width=\columnwidth]{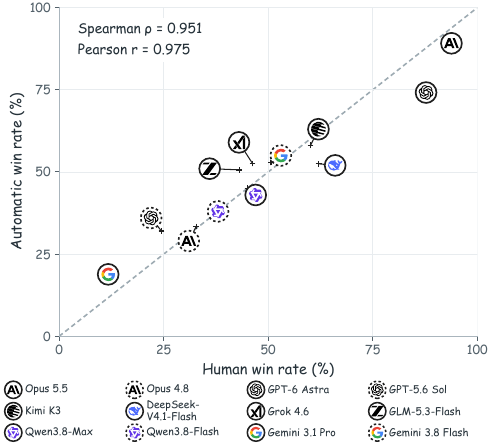}
\caption{\textbf{Overall alignment with human preferences.} Win rates weight the three settings equally within each dimension, then average camera, static, and dynamic consistency equally. Logo badges identify models. Thin leaders locate overlapping results at their exact coordinates. The dashed line denotes equal win rates.}
\label{fig:human_alignment}
\end{figure}

\begin{figure*}[!t]
\centering
\includegraphics[width=\textwidth]{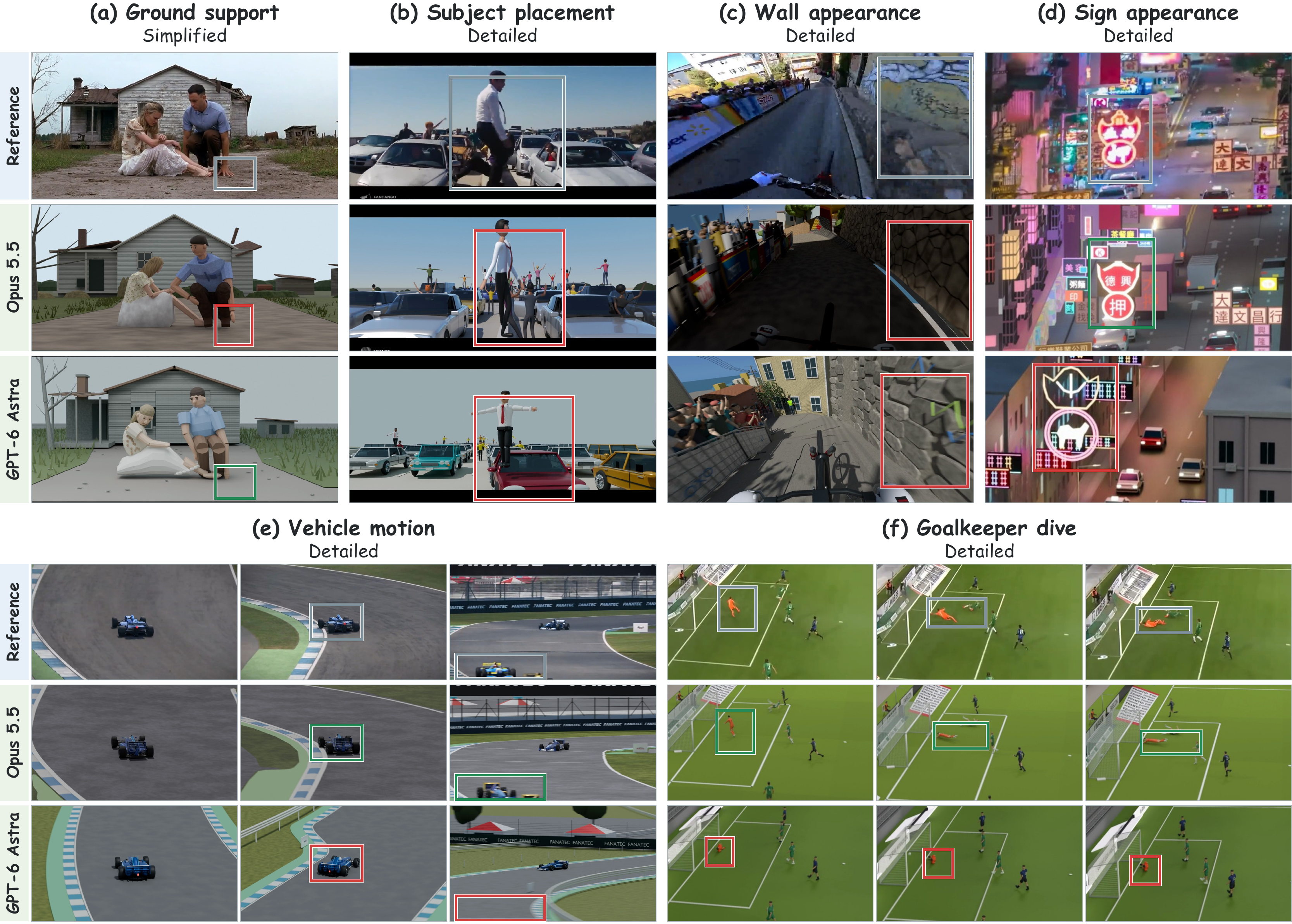}
\caption{\textbf{Spatial, appearance, and motion comparisons.} Each case shows the reference, Opus 5.5, and GPT-6 Astra under the indicated setting. Panels (a)--(d) compare corresponding single frames. Panels (e)--(f) show three-frame sequences progressing from left to right at matched times. Gray boxes locate reference details. Green and red local boxes highlight reproduced details and remaining errors, respectively.}
\label{fig:reconstruction_cases}
\end{figure*}

\begin{figure}[!htb]
\centering
\includegraphics[width=\columnwidth]{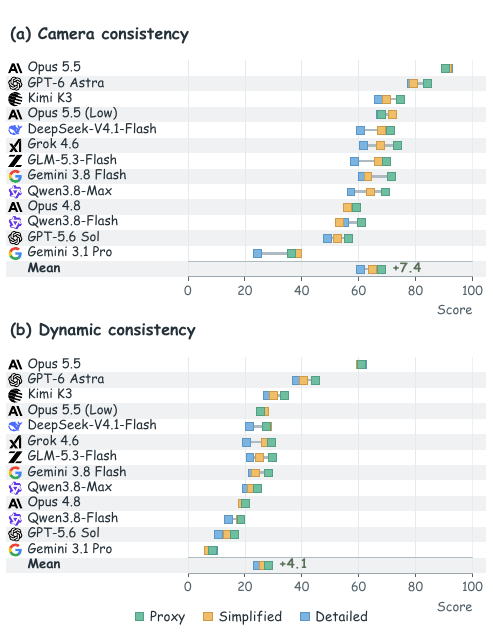}
\caption{\textbf{Representation effects.} Squares show camera and dynamic consistency scores under three settings, with lines connecting each configuration's scores. Scores average checks within clips, then weight clips equally. Mean weights the 13 configurations equally. Annotations report Proxy minus Detailed, computed from unrounded values.}
\label{fig:representation_scores}
\end{figure}

\begin{figure*}[!t]
\centering
\includegraphics[width=\textwidth]{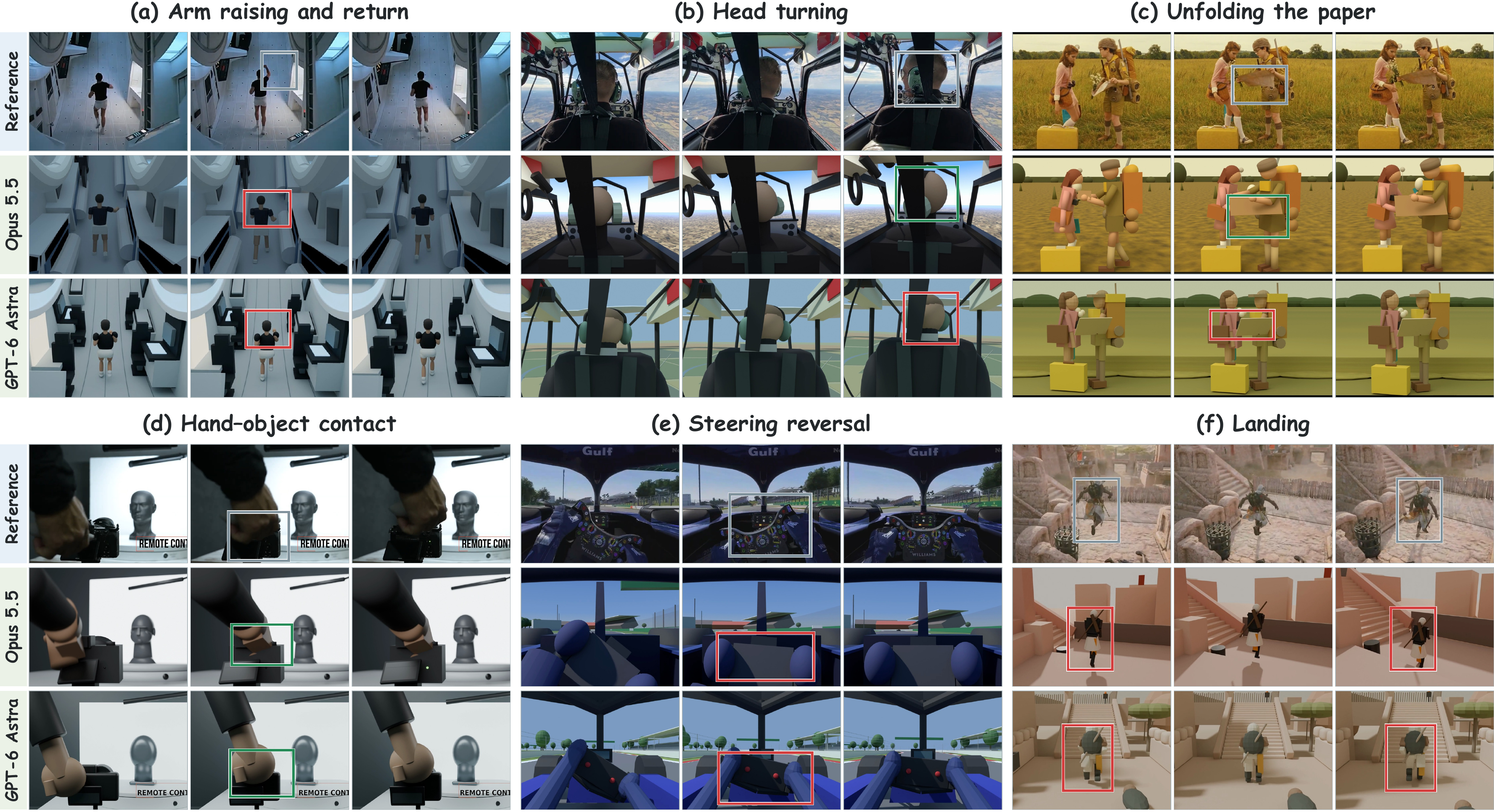}
\caption{\textbf{Action reproduction under Proxy.} Six cases compare the reference, Opus 5.5, and GPT-6 Astra, with three time-matched frames progressing from left to right in each row. The examples cover articulated actions, hand--object interaction, steering, and landing. Gray boxes locate reference details. Green and red local boxes highlight reproduced details and remaining errors, respectively.}
\label{fig:proxy_failures}
\end{figure*}

\begin{figure}[!htb]
\centering
\includegraphics[width=\columnwidth,height=.35\textheight,keepaspectratio]{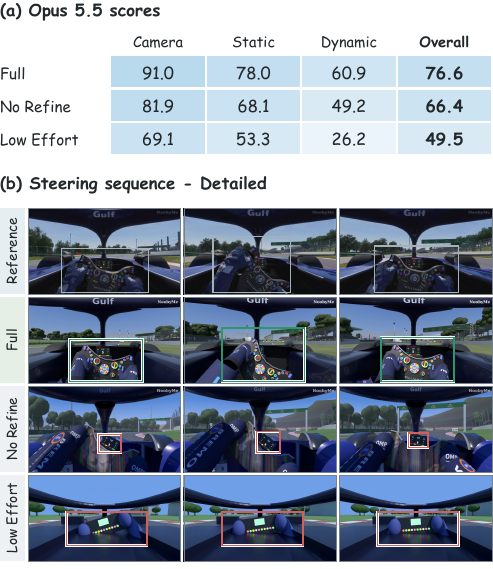}
\caption{\textbf{Iteration and effort.} (a) Opus 5.5 scores under three execution configurations, with equal weights across settings and consistency dimensions. (b) Corresponding frames from a Detailed steering sequence, ordered left to right. Gray boxes mark reference details, green boxes reproduced steering, and red boxes the rotating display (No Refine) or absent steering (Low Effort).}
\label{fig:iteration_effort}
\end{figure}

\section{Analysis}
\label{sec:analysis}

\subsection{Alignment with Human Judgments}
\label{sec:human_alignment}

The results show strong alignment between automatic evaluation and human preferences at the model level (Figure~\ref{fig:human_alignment}). Overall win rates yield a Spearman correlation of 0.951 and a Pearson correlation of 0.975 across the 12 models. Human and automatic evaluation both rank Opus 5.5 and GPT-6 Astra first and second, with three models shared between their top four. Dimension-specific results are provided in Appendix~\ref{sec:appendix_human_evaluation}.

\subsection{Performance and Limitations}
\label{sec:constraint_gaps}

Existing models share relative weaknesses in specific actions, spatial relations, and camera changes. Across all three settings, each of the 13 configurations scores higher on subject motion than action events, object presence than spatial relations, and camera targets than rotation. Figure~\ref{fig:category_scores} presents category scores averaged across representation settings. Scores average items within each clip and category, then weight clips equally for each configuration. The highest model scores on action events are only 42.7, 46.6, and 48.6 in Proxy, Simplified, and Detailed, respectively, indicating that even stronger models incompletely reproduce specific actions and interactions. Category score gaps appear in Appendix~\ref{sec:category_supplement}.

Figure~\ref{fig:reconstruction_cases} compares Opus 5.5 and GPT-6 Astra on the same reference content. Both models miss aspects of the wall's weathered plaster, graffiti, and stone base in (c). This example illustrates shared difficulties in recovering instance-specific appearance, even when the main subjects and scene elements are recognizable.

The temporal comparisons reveal substantial differences in event reproduction. In Figure~\ref{fig:reconstruction_cases}(e), Opus 5.5 follows the car's approach to the curb and includes the foreground vehicles after the cut. GPT-6 Astra deviates from the reference trajectory and omits those vehicles. In (f), Opus 5.5 reproduces the goalkeeper's transition from standing to diving and landing, while GPT-6 Astra places the goalkeeper low near the goal from the first sampled frame. Local contact can show the opposite pattern: in (a), GPT-6 Astra reproduces the ground-support contact that Opus 5.5 misses.

Differences across models are uneven across requirements. In Detailed, the median score across configurations on camera rotation is 25.2 and the highest is 74.9. The corresponding subject-motion scores are 27.3 and 70.3. Comparing Opus 5.5 with GPT-6 Astra, object-presence scores are 89.9 versus 79.6, a gap of 10.3 points. Subject-motion scores are 70.3 versus 44.5 and camera-rotation scores are 74.9 versus 42.1, yielding larger gaps of 25.8 and 32.9 points. Action-event scores are 48.6 versus 27.2. Opus 5.5's larger advantages occur in the reproduction of motion, actions, and viewpoint changes.

\subsection{Representation Effects}
\label{sec:representation_effect}

We compare camera and dynamic consistency across representation settings because the requirements in these two dimensions remain consistent across all three settings. Figure~\ref{fig:representation_scores} shows higher Proxy than Detailed scores for 12 of 13 configurations on camera consistency and 10 of 13 on dynamic consistency. Averaging equally across the 13 configurations, camera and dynamic scores increase by 7.4 and 4.1 points, respectively.

GPT-6 Astra scores 78.5, 79.1, and 84.1 on camera consistency in Detailed, Simplified, and Proxy, respectively. Its dynamic scores are 38.1, 40.6, and 44.6. Opus 5.5 changes little across settings: its Proxy-minus-Detailed differences are $-0.8$ for camera and $-0.5$ for dynamics.

Improved scores still leave dynamic requirements incompletely satisfied. GPT-6 Astra scores 44.6 on dynamic consistency in Proxy and Opus 5.5 scores 60.8. Figure~\ref{fig:proxy_failures}(b) shows that Opus 5.5 reproduces the turn toward the side window, whereas GPT-6 Astra retains a broadly stable head orientation. Both models, however, miss the steering reversal in (e) and the transition from airborne to landing posture in (f). Under Proxy, models can reproduce some articulated actions but still make errors in their direction, timing, and phase transitions.

\subsection{Do Iteration and Effort Help?}
\label{sec:iteration_effort}

We compare three Opus 5.5 configurations. Full (Opus 5.5) uses high reasoning effort and iterative refinement. No Refine retains high effort but omits post-render revision. Low Effort retains iteration with low effort and corresponds to Opus 5.5 (Low) in the main comparisons. Figure~\ref{fig:iteration_effort}(a) reports scores averaged equally across representation settings. Full achieves 76.6 overall, compared with 66.4 for No Refine and 49.5 for Low Effort, gaps of 10.2 and 27.1 points. Both variants score lower across all three dimensions, with the largest reductions in dynamic consistency: 11.6 and 34.6 points.\looseness=-1

In the steering sequence (Figure~\ref{fig:iteration_effort}(b)), Full turns the wheel further and then back. No Refine rotates the display with the wheel. The reference display stays fixed. Low Effort leaves the wheel nearly unchanged across all three instants. Both variants reproduce the wheel's buttons and fine components less faithfully than Full. Iterative refinement and higher reasoning effort improve both dynamic fidelity and the reproduction of visual details.

\section{Related Work}
\label{sec:related_work}

\paragraph{Agentic reconstruction.}
Language models infer structured scene representations from images~\cite{kulits2024re}, while graphics agents edit and reconstruct Blender scenes through visual feedback~\cite{huang2024blenderalchemy,he2026thinking,yin2026vision}.
Code-based generation also supports dynamic scenes from text~\cite{zhang2026code2worlds,liu2026simworlds}.
IR3D-Bench~\cite{liu2026ir3d} evaluates scene reconstruction from images, while 3DCodeBench~\cite{gao20263dcodebench} evaluates procedural object modeling from text and image references.
For video reconstruction, BVB~\cite{tang2026bvb} focuses on indoor scenes with moving cameras, while 4DCodeBench~\cite{shen20264dcodebench} studies physical dynamics in real and synthetic videos with minimal camera motion.
SceneActBench supplies assets and a reference camera for synthetic dynamic reconstruction~\cite{zhao2026sceneactbench}. VisPhyWorld uses two keyframes from synthetic physical scenes under fixed camera configurations~\cite{liang2026visphyworld}.
CineRecon targets filmmaking, evaluating joint reconstruction of scene content, subject actions and interactions, camera movements, and shot transitions.

\paragraph{3D-guided video generation.}
Explicit scene structure and dynamics provide controllable inputs to video generation: rendered scenes supply intermediate guidance~\cite{hu2024scenecraft}, editable meshes control video editing~\cite{liu2025shape}, and global 3D proxies guide camera trajectories~\cite{fang2026spatialcrafter}.
Physical simulation can also specify motion for generative video refinement~\cite{liu2024physgen}.
CineRecon evaluates the upstream reconstruction of editable dynamic scenes from reference videos.

\paragraph{Reconstruction evaluation.}
Video evaluation spans overall quality~\cite{huang2024vbench}, perceptual motion quality~\cite{ling2025vmbench}, and temporal compositionality~\cite{feng2024tc}. Contrastive video--language evaluation probes entity, action, and event-order mismatches~\cite{bansal2024videocon}.
CameraBench~\cite{lin2026towards} examines geometric and semantic aspects of camera motion.
Question-based image evaluation decomposes semantic fidelity into individual checks~\cite{hu2023tifa,cho2024davidsonian}.
BVB and 4DCodeBench~\cite{tang2026bvb,shen20264dcodebench} assess reference fidelity through visual metrics and scene-specific questions.
CineRecon uses check items tied to entities and times to assess camera, static, and dynamic consistency.

\section{Conclusion}
\label{sec:conclusion}

CineRecon benchmarks reconstruction of dynamic cinematic videos as editable Blender scenes through camera, static, and dynamic checks under three representation settings. We evaluate 12 multimodal coding models using their native agent harnesses. Our analysis identifies spatial relations, specific action events, and camera changes as shared weaknesses of these models. Simplifying appearance requirements improves camera and dynamic consistency for most models, yet errors in action timing, contact relations, and coordinated subject motion persist under Proxy. These findings make the joint reconstruction of geometry, actions, and camera behavior a priority for graphics agents. CineRecon provides a basis for measuring progress toward editable visual proxies for controllable video generation. Evaluating their effect on downstream generation is a next step.

\clearpage

\makeatletter
\global\@colroom=\@colht
\global\vsize=\@colht
\makeatother

{
    \small
    \bibliographystyle{ieeenat_fullname}
    \bibliography{references}
}

\clearpage
\onecolumn
\appendix
\setlength{\intextsep}{12pt plus 2pt minus 2pt}
\setlength{\textfloatsep}{14pt plus 2pt minus 2pt}
\setlength{\floatsep}{12pt plus 2pt minus 2pt}
\captionsetup[table]{position=top,skip=6pt}

\raggedbottom
\section{Dataset and Annotation}
\label{sec:appendix_dataset}

\subsection{Dataset Statistics}
\label{sec:dataset_statistics}

Table~\ref{tab:data_collection} summarizes all 50 reference videos and their annotations. The videos total 763.59 seconds, with a mean duration of 15.27 seconds and a range of 4.90--24.04 seconds. The 360 evaluation clips span all reference videos, with 3--13 clips per video. Clips average 1.41 seconds and last at most two seconds. Clip duration is computed as the inclusive frame count divided by the frame rate.

Proxy, Simplified, and Detailed contain 9,647, 20,021, and 20,000 checks, respectively, totaling 49,668 setting-specific checks. A check applicable to multiple settings is counted in each setting. Merging shared records yields 35,836 records in the released checklists. These counts cover the full annotation set and are independent of the number of evaluated models or repeated evaluations. Camera and dynamic checks cover the same categories, subjects, and temporal intervals across settings, with 2,376 camera checks and 703 dynamic checks per setting.

\begin{table}[htbp]
\centering\small
\caption{CineRecon dataset statistics. Videos and evaluation clips are counted once. Check items are counted separately for each applicable representation setting.}
\label{tab:data_collection}
\setlength{\tabcolsep}{12pt}\renewcommand{\arraystretch}{1.12}
\begin{tabular}{@{}lr@{}}\toprule
Statistic & Value \\ \midrule
Reference videos & 50 \\
Total reference duration (s) & 763.59 \\
Mean reference duration (s) & 15.27 \\
Reference duration range (s) & 4.90--24.04 \\
Evaluation clips & 360 \\
Clips per video, range & 3--13 \\
Mean clip duration (s) & 1.41 \\
Clip duration range (s) & 0.42--2.00 \\
Evaluation dimensions / check-item categories & 3 / 11 \\
\midrule
Check items: Proxy & 9,647 \\
Check items: Simplified & 20,021 \\
Check items: Detailed & 20,000 \\
Total check items (counted by setting) & 49,668 \\
\bottomrule\end{tabular}
\end{table}

\subsection{Evaluation Clip and Checklist Construction}
\label{sec:clip_checklist_construction}

We organize reconstruction tasks by complete reference video and annotate local clips for evaluation. Clips cover observable scene states, subject actions and interactions, camera changes, and shot transitions. They may overlap or cross shot boundaries. All clips and check items were initially annotated by Codex powered by GPT-6 Astra, followed by multiple rounds of human review and refinement. Review used reference frames and temporal sequences to check subject correspondence, visible facts, temporal ranges, categories, and applicable representation settings, and to revise statements inconsistent with or insufficiently supported by the visual evidence.

Each check item records its category, applicable settings, referenced subjects, local start and end frames, and an assessable requirement. Frame-specific checks describe visible states, such as subject counts, relative positions, or hand-ground contact. Interval checks describe temporal changes, such as camera translation, subject trajectories, or a crouching action. Items are organized into the eleven categories defined in the main text, separating camera behavior, static states, and dynamic processes. Action and interaction statements specify observable progression and timing. Appearance statements concern structures and attributes supported by the frames.

Representation settings determine the granularity of appearance checks. Proxy assesses object presence, spatial relations, camera behavior, and dynamics. Simplified appearance checks require recognizable categories, overall silhouettes, major structures, and salient attributes. Detailed additionally requires visible local structures and details of the reference instance. Camera and dynamic checks retain the same categories, subjects, and temporal intervals across settings, supporting comparisons of these capabilities as appearance requirements are reduced.

\subsection{Subject Identification and Temporal Annotation}
\label{sec:subject_temporal_annotation}

Entity tables assign fixed names to individually tracked people, animals, and vehicles, such as Person A, Person B, and Vehicle A. Interaction objects may also receive identifiers when similar objects need to be distinguished. Scene regions such as ground and grass retain descriptive names. Group entries specify their scope to avoid counting individually identified members twice. Subjects with established correspondence retain stable identifiers within a source video, and the three settings share the same entity table for each clip. Subsequent checks use these fixed names, with appearance attributes stated as separate requirements in the applicable items.

Each numbered subject has a short identifying description and a first-appearance frame. First appearance denotes the earliest identifiable visible part that can be linked to the subject through the temporal sequence. Color and clothing descriptions in the entity table locate the reference subject, while check statements and their applicable settings define the scoring requirements. Local frame indices start at zero, and temporal intervals include both endpoints. The released records also store the clip's frame offset in the source video, mapping local annotations to the complete reference. Spatial statements specify image directions or relative reference objects, and subject correspondence remains consistent throughout each annotated motion or interaction interval.

\subsection{Annotation Example}
\label{sec:annotation_example}

Clip \texttt{video38\_cp006} spans local frames 0--46, corresponding to source frames 256--302. At approximately 23.976 frames per second, it lasts 1.96 seconds. Person A sits on a dirt road in front of a wooden house, while Person B crouches on A's image-right and gradually places a hand on the ground for support. Figure~\ref{fig:annotation_sequence} compares the reference with Opus 5.5's Simplified reconstruction. Table~\ref{tab:annotation_entities} lists the two numbered subjects. The full clip checklist contains 55 checks for this setting.

\noindent\begin{minipage}{\linewidth}
\centering
\includegraphics[width=\linewidth]{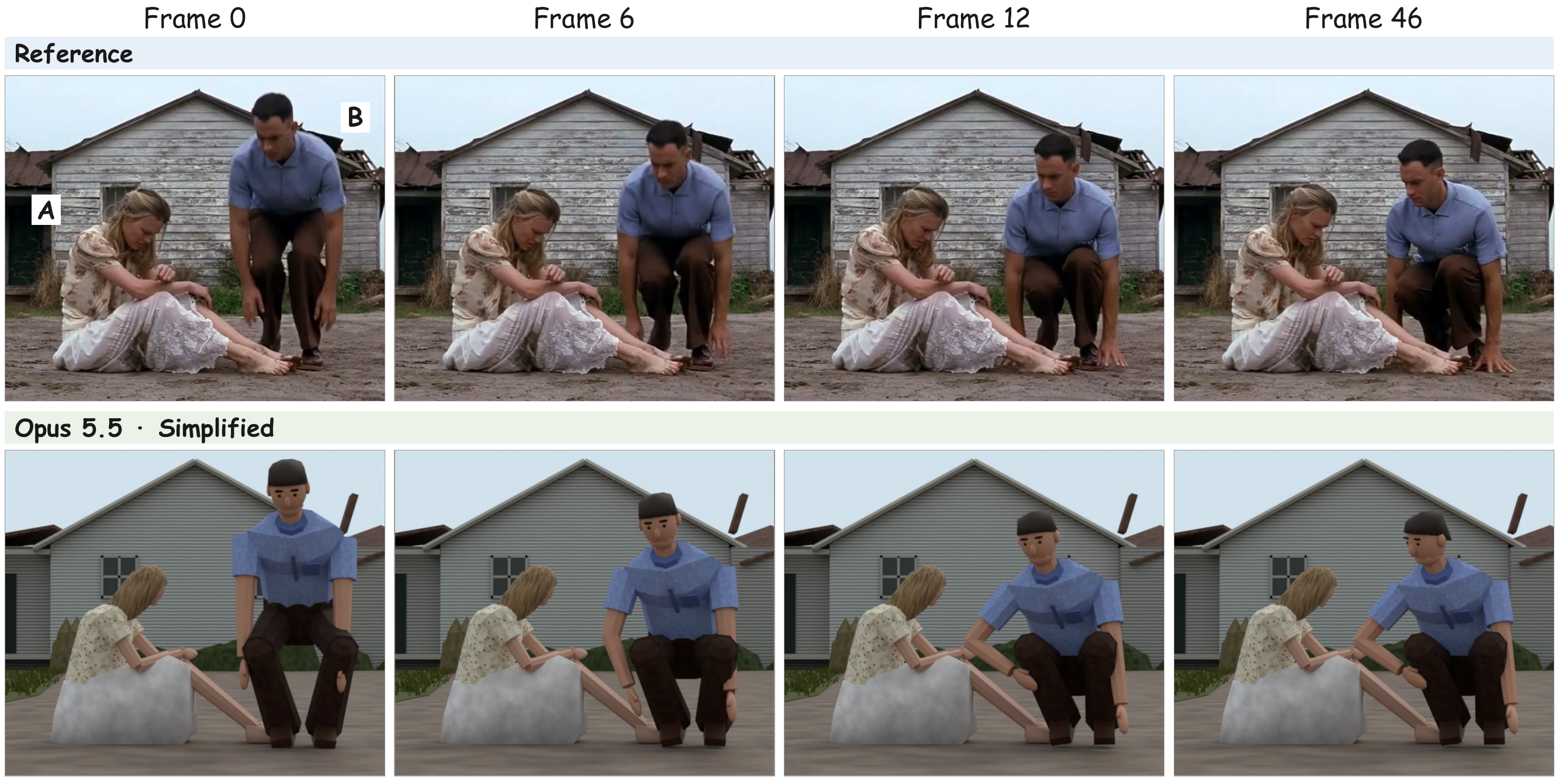}
\captionof{figure}{\textbf{Annotation and scoring example.} The reference (top) and Opus 5.5's Simplified reconstruction (bottom) are aligned at local frames 0, 6, 12, and 46. A and B identify the subjects in Table~\ref{tab:annotation_entities}. The reconstruction reproduces a crouching pose, but the hand does not establish supporting contact with the ground in the final frame. Corresponding checks and VLM judge ratings appear in Table~\ref{tab:annotation_checks}. Both rows use the same crop. Temporal samples are not equally spaced.}
\label{fig:annotation_sequence}
\end{minipage}

Table~\ref{tab:annotation_checks} presents eight Simplified checks and their VLM judge ratings. The presence of both subjects, Person B's main clothing, the initial spatial configuration, and the fixed-camera requirements are rated fully satisfied. The final hand-support relation and the action requirement to establish and maintain support after crouching are rated unsatisfied. Both failures concern hand-ground contact, showing how recognizable subjects and a crouching pose can still omit the required support relation.

\begin{table}[htbp]
\centering\small
\caption{The two numbered subjects in the example clip. First appearances use local frame indices.}\label{tab:annotation_entities}
\setlength{\tabcolsep}{4pt}\renewcommand{\arraystretch}{1.15}
\begin{tabular}{@{}p{.13\linewidth}p{.15\linewidth}p{.66\linewidth}@{}}\toprule
Name & First frame & Identification \\ \midrule
Person A & 0 & Long light-colored hair, a beige patterned short-sleeved top, and a long white skirt; seated barefoot on the dirt road, slightly left of center. \\ \addlinespace
Person B & 0 & Short dark hair, a light-blue short-sleeved collared shirt, brown trousers, and dark-brown shoes; leaning forward and crouching to the image-right of Person A. \\
\bottomrule\end{tabular}
\end{table}

\medskip

\begin{table}[htbp]
\centering\small
\caption{Selected Simplified checks and VLM judge ratings for Opus 5.5 on this clip. $\checkmark$: fully satisfied (1); $\times$: unsatisfied (0). None of these eight checks received a partially satisfied (0.5) rating. Statements are translated from the released checklist.}\label{tab:annotation_checks}
\setlength{\tabcolsep}{4pt}\renewcommand{\arraystretch}{1.13}
\begin{tabular}{@{}p{.14\linewidth}p{.08\linewidth}p{.64\linewidth}c@{}}\toprule
Category & Frames & Check item & Rating \\ \midrule
Object presence & 0 & Person A is clearly visible. & \textcolor{green!45!black}{$\checkmark$} \\
\addlinespace
Object presence & 0 & Person B is clearly visible. & \textcolor{green!45!black}{$\checkmark$} \\
\addlinespace
Object appearance & 0 & Person B has short dark hair and wears a light-blue short-sleeved collared shirt, brown trousers, and dark-brown shoes. & \textcolor{green!45!black}{$\checkmark$} \\
\addlinespace
Spatial relations & 0 & Person B is to the image-right of Person A, beside A's outstretched feet, with knees bent and torso leaning forward; both hands remain above the dirt road. & \textcolor{green!45!black}{$\checkmark$} \\
\addlinespace
Spatial relations & 46 & Person B crouches low to the image-right of Person A, beside A's outstretched feet, with the hand on the image-right supporting against the dirt road. & \textcolor{red!70!black}{$\times$} \\
\addlinespace
Camera translation & 0--46 & The camera remains at a fixed position, with no visible horizontal or vertical translation. & \textcolor{green!45!black}{$\checkmark$} \\
\addlinespace
Shot transitions & 0--46 & The clip is a continuous shot with zero shot transitions. & \textcolor{green!45!black}{$\checkmark$} \\
\addlinespace
Action events & 0--46 & To the image-right of Person A, Person B continues bending the knees from a forward-leaning pose and lowers into a crouch. The hand on the image-right gradually reaches the dirt road and establishes support; the low crouch and hand-ground contact persist in the later part of the clip. & \textcolor{red!70!black}{$\times$} \\
\bottomrule\end{tabular}
\end{table}

\section{Reconstruction Task and Configuration}
\label{sec:appendix_configuration}

\subsection{Task Prompts for the Three Representation Settings}
\label{sec:task_prompts}

Each reconstruction task supplies the complete reference video, a representation-specific instruction, and an output directory. The agent selects the frames and temporal intervals used for its own analysis and verification. Evaluation clips, entity tables, checklists, and reference camera parameters are withheld. The following summarizes the shared instructions and setting-specific requirements of the task prompts.

\begin{promptbox}{Reconstruction prompt: shared instructions (summary)}
Reconstruct the reference video at \texttt{<VIDEO\_PATH>} as an editable dynamic scene in Blender 5.2, and save the deliverables to \texttt{<OUTPUT\_DIR>}. Analyze the video, construct and animate the scene, inspect rendered feedback, and revise the reconstruction autonomously. Preserve subject count and correspondence, relative size and position, orientation, occlusion, support and contact, motion paths, articulated actions, interactions, and event order. Reproduce camera framing, perspective, movement, and shot transitions at the corresponding times. Match the reference resolution, aspect ratio, frame rate, frame count, and duration.

Inspect the scene with Blender MCP before construction, after substantial edits, and before submission. Compare reference and rendered images at matching times across the shots and key events; revise discrepancies within the assigned representation setting. Use editable 3D objects, and render the final video with Cycles on a CUDA GPU with denoising.
\end{promptbox}

\begin{promptbox}{Representation-specific requirements (summary)}
\textbf{Proxy.} Use a small number of primitives or simple combinations to preserve subject count, coarse proportions, relative size, placement, and orientation. Separate moving parts when required for an articulated action or interaction. Category-specific detailed geometry, facial features, detailed clothing, textures, and decoration are unnecessary. 

\textbf{Simplified.} Construct recognizable low-complexity objects with the main silhouette, proportions, and important components. Approximate body shape, clothing or structural features, major colors, salient patterns, and material categories while preserving the reference style. Fine geometry, detailed textures, minor decoration, and subtle secondary motion may be simplified. Articulation and contact must remain sufficient to reproduce the observed events.

\textbf{Detailed.} Reproduce the visible global and local shape, component structure, instance-specific appearance, and motion of the reference subjects. Preserve the original visual style and visible imaging effects, including lighting, shadows, reflections, transparency, depth of field, and motion blur. Verification covers these visible details together with scene layout, camera behavior, action timing, and contact. The prompt instructs the agent to continue correcting observable discrepancies found during its checks.
\end{promptbox}

The requested deliverables are \texttt{scene.blend}, \texttt{reconstruction.mp4}, a synchronized side-by-side comparison video, a comparison contact sheet, and a short implementation and verification report. Reference footage appears on the left and reconstruction on the right in the comparison. Each rendered frame must correspond to the same time in the reference. Temporal shifts, speed changes, interpolation, and repeated frames must not conceal reconstruction errors.

\subsection{Agent Tools and Execution Environment}
\label{sec:agent_environment}

The reconstruction environment uses Blender 5.2~\cite{blender2026} with Blender MCP~\cite{ahuja2025blendermcp}, alongside the file, shell, and media-inspection tools provided by each agent harness. Blender MCP exposes scene inspection, Python execution within Blender, and viewport capture. The agent can construct geometry and materials, configure lights and cameras, animate objects and articulated parts, inspect scene state, and save the editable project. Local Python utilities and FFmpeg/FFprobe support frame extraction, image comparison, video encoding, and checks of the output timeline. The agent determines its analysis frames, scene parameterization, animation implementation, and intermediate render settings.

Tasks run on Linux GPU workers. A graphical Blender process runs under Xvfb to support the MCP connection and viewport inspection. The prompts request CUDA rendering with denoising and permit reduced-resolution test renders. Final outputs must retain the reference resolution and timeline. Sampling and bounce settings are selected during reconstruction. Rendering-performance settings are recorded with the run.

The reconstruction loop consists of reference inspection, scene editing, test rendering, visual comparison, and further editing. The agent chooses the number and location of intermediate checks. Its feedback consists of reference media, rendered images, and tool outputs. Checklist ratings from the VLM judge are produced in the separate evaluation stage. Before submission, the prompt requests a final scene inspection, nonempty deliverables, a decodable video, and a check of its timeline.

\subsection{Model Configuration}
\label{sec:model_configuration}

Table~\ref{tab:agent_configuration} summarizes the agent harnesses and reasoning settings for reconstruction. Each model is evaluated through its corresponding harness, which supplies the interaction loop, context handling, and native tools. The three representation settings share the corresponding model's agent harness. The quantitative comparisons include 13 configurations: xhigh for the two Qwen models, high for the other ten models, and low for Opus 5.5, with a budget of 32,768 output tokens per model response. The Opus 5.5 ablation also includes No Refine, as specified in Sec.~\ref{sec:iteration_effort}.

\begin{table}[htbp]
\centering
\caption{\textbf{Agent harnesses and reasoning settings.}}
\label{tab:agent_configuration}
\small\setlength{\tabcolsep}{5pt}
\begin{tabular}{@{}lll@{}}
\toprule
Model & Agent harness & Reasoning \\
\midrule
GPT-6 Astra~\cite{openai2026astra} & Codex CLI 0.153.4 & high \\
GPT-5.6 Sol~\cite{openai2026sol} & Codex CLI 0.153.4 & high \\
Opus 5.5~\cite{anthropic2026opus55} & Claude Code 2.1.280 & high/low \\
Opus 4.8~\cite{anthropic2026opus48} & Claude Code 2.1.278 & high \\
GLM-5.3-Flash~\cite{zai2026glm53} & ZCode CLI 0.16.9 & high \\
Qwen3.8-Max~\cite{alibaba2026qwenmax} & Qwen Code 0.24.1 & xhigh \\
Qwen3.8-Flash~\cite{alibaba2026qwenflash} & Qwen Code 0.24.1 & xhigh \\
Kimi K3~\cite{moonshot2026kimik3} & Kimi Code CLI 2.0.2 & high \\
Grok 4.6~\cite{xai2026grok46} & Grok Build 1.0.41 & high \\
DeepSeek-V4.1-Flash~\cite{deepseek2026v41} & DSH 0.1.5-rc.3 & high \\
Gemini 3.1 Pro~\cite{google2026gemini31} & Antigravity CLI 1.2.11 & high \\
Gemini 3.8 Flash~\cite{google2026gemini38} & Antigravity CLI 1.2.11 & high \\
\bottomrule
\end{tabular}
\end{table}

The task prompts prescribe iterative checking and correction without a fixed number of refinement rounds. A complete reconstruction includes multiple model calls, tool executions, and renders. These instructions apply to the full reconstruction setting. The No Refine ablation omits iterative refinement, as described in Sec.~\ref{sec:iteration_effort}.

\section{Automatic Evaluation Details}
\label{sec:appendix_automatic_evaluation}

\subsection{Temporal Alignment of Reference and Reconstructed Videos}
\label{sec:temporal_alignment}

Evaluation uses the complete rendered video and the temporal annotations of each reference clip. Let $o_c$ be the source-frame offset of clip $c$, $f$ a zero-based local frame index, and $r$ the reference frame rate. The corresponding time in the complete reference video is
\begin{equation}
 t_c(f)=\frac{o_c+f}{r}.
\end{equation}
A check on local frames $[a,b]$ includes both endpoints and covers the interval $[(o_c+a)/r,(o_c+b+1)/r)$. Reference and reconstruction share the same time origin. At the required frame rate, local frame $f$ corresponds to rendered frame $o_c+f$. When the encoded frame rate differs, the pipeline locates the generated interval from the reference clip's start time and selects generated frames nearest to the requested local times.

Frame selection preserves the evidence needed by the checklist. Mandatory frames include the first and last frames of the clip, the annotated evidence frames and endpoints of each applicable check, and the frames immediately before, at, and after each annotated shot transition, where these fall within the clip. Additional candidates include interval midpoints, annotation-specified sample frames, and samples at approximately eight frames per second. A request contains at most 50 aligned frame pairs. When the candidate set exceeds this limit, all mandatory frames are retained. The remaining slots are filled by successively choosing the candidate farthest from the selected frames, with earlier frames breaking ties. This maintains coverage between the required observations. The released annotations fit within this limit for all three settings.

Each selected time is presented as one vertically stacked image, with the reference above and the reconstruction below. Both images retain their aspect ratio within a $960\times540$-pixel area, and labels indicate their roles, local frame index, and local timestamp. The pairs are provided in increasing time order. The judge therefore receives individual aligned observations together with the temporal sequence needed to assess camera changes, motion, and interactions. Before evaluation, the pipeline checks video decoding, availability of the complete requested interval, and consistency between the saved frames and the source files.

\subsection{VLM Judge Prompts and Rating Rules}
\label{sec:judge_prompts}

We use GPT-5.6 Sol~\cite{openai2026sol} as the VLM judge with high reasoning effort. One request evaluates one reconstruction clip under one representation setting. It contains the setting-specific judging instruction, the aligned frame pairs, a reference entity table, and the applicable checklist. The entity table supplies stable identifiers, names, identification descriptions, and first-appearance frames. Each check specifies its statement, category, subject identifiers, local temporal range, and evidence frames. All checks for the clip are judged in the same request.

The shared judging instruction can be summarized as follows:
\begin{promptbox}{VLM judge prompt: shared instructions (summary)}
Treat SOURCE as the reference and GENERATED as the reconstruction. Establish subject correspondence using reference identity cues and temporal continuity, then preserve that correspondence throughout the clip. Judge each statement only at its specified local frame or interval; frame indices start at zero and interval endpoints are inclusive. Assess camera changes within shots, keeping shot transitions separate from continuous camera motion. Check whole-subject motion and action events according to their respective statements. Return exactly one rating per item, in the original item order.
\end{promptbox}

Subject names and identification descriptions establish which reference entity a check concerns. The judge maintains the same reference-to-reconstruction correspondence across the checks and must not exchange subjects to satisfy individual relations. Group members are not counted again as independent instances, and mirror reflections are not counted as additional people. A missing subject or reference object required by a relation makes that relation unsatisfied. Checks that explicitly describe occlusion or leaving the frame are assessed according to their stated conditions.

The setting-specific instructions define which appearance properties enter the judgment. Under Proxy, simple distinguishable geometric stand-ins are allowed. Reference names and appearance descriptions serve as identity cues, and category-specific shape, color, clothing, material, texture, and decoration are not appearance requirements. Object presence, count, spatial relations, camera behavior, motion, and interaction remain subject to the checklist. Under Simplified, appearance checks require recognizable categories, silhouettes, proportions, important components, and the salient colors, patterns, clothing or structures, and material categories explicitly stated in the item. Detailed additionally checks clearly visible local features and reference-specific appearance. Across settings, presence, relation, camera, and dynamic checks are judged by their own statements without adding independent appearance conditions. Allowed simplifications incur no penalty, and neither pixel equality nor unspecified invisible details is required.

Table~\ref{tab:judge_ratings} defines the three ratings and their numerical values. Each item receives one rating for its full statement. The judge returns a JSON object with a single \texttt{results} array whose length and order match the input checklist. The response contains rating labels only. Item identifiers and numerical scores are attached by the evaluation code. Figure~\ref{fig:annotation_sequence} and Table~\ref{tab:annotation_checks} provide a concrete example of the evidence and resulting judgments.

\begin{table}[htbp]
\centering
\caption{\textbf{Check-item rating rules.} Scores reflect agreement with the statement at its specified frame or interval under the assigned representation setting.}
\label{tab:judge_ratings}
\small\setlength{\tabcolsep}{4pt}
\begin{tabular}{@{}p{0.20\textwidth}cp{0.68\textwidth}@{}}
\toprule
Rating & Score & Criterion \\
\midrule
Unsatisfied & 0 & A required subject is absent, a key fact is missing or reversed, or the main required relation does not hold. \\
\addlinespace
Partially satisfied & 0.5 & The main fact or salient feature holds, but some required attributes, spatial relations, timing, motion, or contact remain visibly incorrect. \\
\addlinespace
Fully satisfied & 1 & All observable requirements of the statement hold at the specified frame or interval, allowing the simplifications permitted by the setting. \\
\bottomrule
\end{tabular}
\end{table}

\subsection{Score Aggregation and Failure Handling}
\label{sec:aggregation_failures}

Let $y_{m,s,c,i}\in\{0,0.5,1\}$ denote the score for item $i$ in clip $c$, obtained by model configuration $m$ under setting $s$. Let $\mathcal{I}_{s,c,d}$ be the set of applicable items in consistency dimension $d$. We first average the item scores within that clip and dimension:
\begin{equation}
 q_{m,s,c,d}=\frac{1}{|\mathcal{I}_{s,c,d}|}
 \sum_{i\in\mathcal{I}_{s,c,d}}y_{m,s,c,i}.
\end{equation}
Writing $\mathcal{C}_{s,d}$ for the annotated clips containing at least one item in dimension $d$, the dimension score is
\begin{equation}
 S_{m,s,d}=\frac{100}{|\mathcal{C}_{s,d}|}
 \sum_{c\in\mathcal{C}_{s,d}}q_{m,s,c,d},
 \qquad
 S_{m,s}=\frac{S_{m,s,\mathrm{cam}}+S_{m,s,\mathrm{stat}}+S_{m,s,\mathrm{dyn}}}{3}.
\end{equation}
Scores are reported on a 0--100 scale. A clip with no annotated item in a dimension does not contribute to that dimension's mean. Clips are equally weighted within each dimension. Videos contribute through their annotated clips, without a further video-level averaging step. The three dimension scores receive equal weight because their checklist sizes differ substantially. Category scores follow the same item-within-clip and then clip-average procedure. Each configuration and setting is aggregated independently, and a mean row across configurations gives each configuration equal weight.

Before accepting a judgment, the pipeline verifies that the response is complete, contains the expected JSON fields, and returns exactly one permitted label for every input item. Truncated responses, refusals, invalid labels, and missing or extra ratings are recorded as evaluation failures. Recoverable request failures, including timeouts and rate limits, trigger retries with the same evidence and checklist. A score is attached only after the response passes these checks. The saved record links each item to its label, numerical score, input evidence, and judge configuration.

Visual failures and execution failures have separate records. A missing subject, incorrect contact, or absent action in a decodable reconstruction receives the corresponding checklist rating. A missing or undecodable video, an unavailable temporal interval, or a failed judge request is recorded as an operational failure without assigning a visual score of zero. The pipeline retries the failed operations and records scores only after obtaining usable video evidence and a valid judge response.

\section{Human Evaluation}
\label{sec:appendix_human_evaluation}

\subsection{Sample Selection and Comparison Setup}
\label{sec:human_samples_interface}

Human evaluation uses dimension-specific pairwise comparisons to assess whether relative model performance under automatic evaluation agrees with human preferences. Each task presents a reference clip, reconstructions from two models under the same representation setting, and one evaluation dimension. The study contains 1,440 tasks covering 50 reference videos and 12 models, with 480 tasks each for camera, static, and dynamic consistency. Reasoning settings follow Table~\ref{tab:agent_configuration}. Table~\ref{tab:human_sample_counts} summarizes the allocation across representation settings. Comparisons of the same reconstruction pair on different dimensions count as separate tasks.

\begin{table}[htbp]
\centering\small
\setlength{\tabcolsep}{4pt}\renewcommand{\arraystretch}{1.2}
\caption{Human comparison tasks by dimension and representation setting.}
\label{tab:human_sample_counts}
\begin{tabular}{@{}lrrrr@{}}\toprule
Dimension & Proxy & Simplified & Detailed & Total \\ \midrule
Camera & 161 & 160 & 159 & 480 \\
Static & 159 & 161 & 160 & 480 \\
Dynamic & 160 & 159 & 161 & 480 \\ \midrule
Total & 480 & 480 & 480 & 1,440 \\
\bottomrule\end{tabular}
\end{table}

Sampling uses exposure counts for videos, settings, models, and model pairs. Score magnitudes and differences do not enter selection. Camera and static comparisons each cover all 50 videos, with nine or ten tasks per video. Dynamic comparisons draw from the 47 videos with applicable dynamic checks, with nine to eleven tasks per video. Within each dimension, each model appears 79--81 times, and each of the 66 unordered model pairs appears seven or eight times. This allocation provides similar exposure across videos, models, and opponents.

Camera and dynamic consistency are judged from temporally aligned reference and reconstruction clips. Static consistency is judged from reference and reconstructed frames at times specified by applicable checks. The candidates are anonymized as A and B. Raters receive the current dimension and representation requirements, while model names, execution configurations, and automatic scores are withheld. Task order is independently randomized for each rater, and assignment to A or B is balanced within each dimension.

\subsection{Annotation Procedure and Preference Aggregation}
\label{sec:human_preference_procedure}

Raters compare fidelity to the reference along the specified dimension. Camera judgments concern viewpoint, framing, camera motion, and shot transitions. Static judgments concern object presence, structure, appearance, and spatial relations. Dynamic judgments concern motion directions and paths, specific actions, interactions, and their timing. Representation requirements follow the main protocol: Proxy permits geometric stand-ins, Simplified requires recognizable structures and salient appearance, and Detailed requires reference-specific visible details. All settings retain the corresponding spatial, camera, and dynamic requirements.

Three raters independently assess all 1,440 comparison tasks. For each task, they choose whether A better matches the reference, the two results show no clear difference, or B better matches the reference. They cannot see one another's labels before submitting. No clear difference denotes similar reconstruction quality on the specified dimension. Tasks with playback failures, insufficient evidence, or inapplicable dimensions are recorded as issues with an explanation.

Preferences are aggregated by majority vote: an option selected by at least two of the three raters becomes the task label. Tasks with an issue report or three different votes are excluded from both human and automatic aggregation.

\subsection{Agreement with Automatic Evaluation}
\label{sec:human_auto_agreement}

Human preferences and automatic scores are both converted into pairwise win, tie, and loss scores. For the aggregated human preference, a win earns 1, a loss earns 0, and a tie earns 0.5 for each model. Automatic comparison uses the two models' scores for the same dimension, clip, and representation setting. The higher score earns 1 and the lower score earns 0. Exact equality of the original scores yields 0.5 for each model. Both sides compute win rates on the same tasks with valid preference labels.

Let $\mathcal{T}_{m,s,d}$ denote the valid comparison tasks involving model $m$ in setting $s$ and dimension $d$. Let $z^{H}_{m,t}$ and $z^{A}_{m,t}$ be its human and automatic comparison scores in task $t$. Both win rates are computed as
\begin{equation}
 w^{X}_{m,s,d}=\frac{1}{|\mathcal{T}_{m,s,d}|}
 \sum_{t\in\mathcal{T}_{m,s,d}}z^{X}_{m,t},\qquad X\in\{H,A\}.
\end{equation}
The three representation settings then receive equal weight:
\begin{equation}
 \bar{w}^{X}_{m,d}=\frac{1}{3}\sum_s w^{X}_{m,s,d}.
\end{equation}
The overall win rate gives equal weight to the three dimensions:
\begin{equation}
 \bar{w}^{X}_{m}=\frac{1}{3}\sum_d\bar{w}^{X}_{m,d}.
\end{equation}
Win rates count a tie as half a win and are reported as percentages.

For the overall result and each dimension, agreement across the 12 models is measured with Spearman correlation between human and automatic win rates. Pearson correlation additionally describes their numerical relationship. Each point in the corresponding scatter plot represents one model, with human win rate on the horizontal axis and automatic win rate on the vertical axis. This analysis assesses alignment with human judgments at the level of relative model performance.

Of the 1,440 comparison tasks, 1,360 yield valid preference labels: 448 for camera consistency and 456 each for static and dynamic consistency. The excluded tasks comprise 19 issue reports and 61 three-way splits. Table~\ref{tab:human_alignment_results} reports the resulting model-level correlations.

\begin{table}[htbp]
\centering\small
\caption{Human--automatic alignment. Correlations are computed across all 12 models. $N$ counts valid comparison tasks.}
\label{tab:human_alignment_results}
\begin{tabular}{@{}lrrr@{}}\toprule
Dimension & $N$ & Spearman $\rho$ & Pearson $r$ \\ \midrule
Camera & 448 & 0.902 & 0.959 \\
Static & 456 & 0.965 & 0.951 \\
Dynamic & 456 & 0.832 & 0.904 \\
Overall & 1,360 & 0.951 & 0.975 \\
\bottomrule\end{tabular}
\end{table}

Figure~\ref{fig:human_alignment_dimensions} separates the three consistency dimensions. Spearman correlations are 0.902, 0.965, and 0.832 for camera, static, and dynamic consistency. The corresponding Pearson correlations are 0.959, 0.951, and 0.904. Human and automatic evaluation rank Opus 5.5 and GPT-6 Astra first and second in all three dimensions, and identify the same top-four ordering for static consistency.

\noindent\begin{minipage}{\linewidth}\centering
\includegraphics[width=\linewidth]{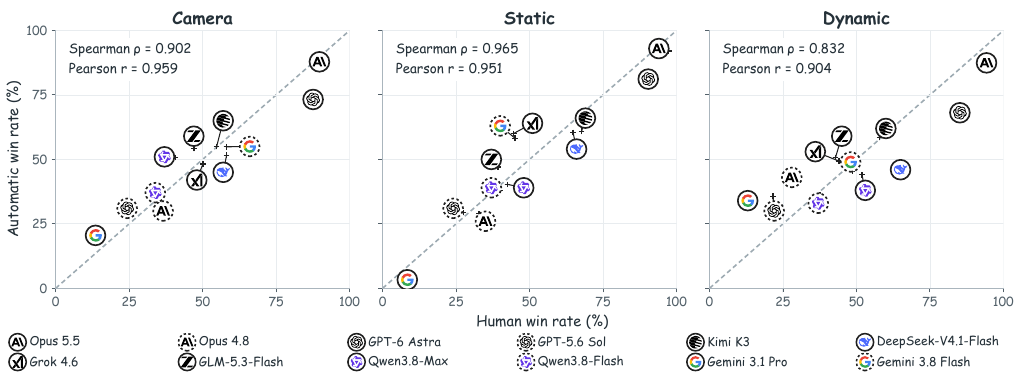}
\captionof{figure}{\textbf{Alignment by consistency dimension.} Human and automatic win rates use the same comparison tasks and weight the three settings equally. Solid and dashed rings distinguish versions from the same provider. Thin leaders locate overlapping results at their exact coordinates. The dashed line denotes equal win rates.}
\label{fig:human_alignment_dimensions}
\end{minipage}

\section{Full Results and Additional Visualizations}
\label{sec:appendix_full_results}

\subsection{Complete Quantitative Results}
\label{sec:full_setting_scores}

Table~\ref{tab:full_scores} reports overall and dimension scores for all 13 configurations under the three representation settings. Each configuration and setting is aggregated independently using Appendix~\ref{sec:aggregation_failures}. The mean row weights configurations equally. Overall scores summarize reconstruction performance, while dimension scores separate camera behavior, scene content, and subject dynamics.

\begin{table}[htbp]
\centering\footnotesize
\setlength{\tabcolsep}{3pt}\renewcommand{\arraystretch}{1.25}
\caption{Scores for 13 model configurations across the three representation settings (0--100). Dimension scores weight clips equally. The overall score averages the three dimensions. Configurations are ordered by their mean overall score across settings. The mean row weights configurations equally. Low denotes low reasoning effort.}
\label{tab:full_scores}
\begin{tabular}{@{}lrrrrrrrrrrrr@{}}\toprule
Model & \multicolumn{4}{c}{Proxy} & \multicolumn{4}{c}{Simplified} & \multicolumn{4}{c}{Detailed} \\
\cmidrule(lr){2-5}\cmidrule(lr){6-9}\cmidrule(lr){10-13}
 & Overall & Cam. & Static & Dyn. & Overall & Cam. & Static & Dyn. & Overall & Cam. & Static & Dyn. \\ \midrule
Opus 5.5 & 76.5 & 90.5 & 78.2 & 60.8 & 76.9 & 91.0 & 79.2 & 60.4 & 76.4 & 91.3 & 76.6 & 61.3 \\
GPT-6 Astra & 67.6 & 84.1 & 74.0 & 44.6 & 63.0 & 79.1 & 69.3 & 40.6 & 60.8 & 78.5 & 65.6 & 38.1 \\
Kimi K3 & 57.3 & 74.6 & 63.2 & 34.0 & 51.8 & 69.7 & 55.8 & 30.1 & 48.2 & 66.8 & 49.7 & 28.0 \\
Opus 5.5 (Low) & 49.1 & 67.9 & 53.9 & 25.4 & 51.4 & 71.9 & 55.6 & 26.7 & 48.1 & 67.5 & 50.3 & 26.5 \\
DeepSeek-V4.1-Flash & 53.9 & 71.0 & 63.3 & 27.6 & 51.1 & 68.0 & 57.3 & 28.0 & 43.4 & 60.6 & 48.0 & 21.7 \\
Grok 4.6 & 54.7 & 73.5 & 61.4 & 29.3 & 49.7 & 67.6 & 54.2 & 27.2 & 42.6 & 61.6 & 45.8 & 20.5 \\
GLM-5.3-Flash & 53.8 & 69.8 & 61.9 & 29.8 & 47.9 & 66.8 & 51.9 & 25.0 & 41.2 & 58.6 & 43.3 & 21.7 \\
Gemini 3.8 Flash & 52.8 & 71.4 & 59.0 & 28.1 & 45.7 & 63.2 & 50.3 & 23.6 & 42.7 & 61.2 & 44.3 & 22.7 \\
Qwen3.8-Max & 49.8 & 69.3 & 56.0 & 24.2 & 45.8 & 64.2 & 51.2 & 22.1 & 40.0 & 57.2 & 42.1 & 20.5 \\
Opus 4.8 & 43.3 & 59.3 & 50.6 & 20.2 & 39.1 & 56.0 & 42.4 & 19.0 & 38.4 & 56.7 & 38.5 & 20.1 \\
Qwen3.8-Flash & 42.7 & 60.9 & 48.7 & 18.6 & 38.5 & 53.3 & 43.7 & 18.5 & 36.5 & 55.1 & 40.5 & 14.0 \\
GPT-5.6 Sol & 39.7 & 56.4 & 46.4 & 16.4 & 36.5 & 52.5 & 43.1 & 13.9 & 31.5 & 48.8 & 35.2 & 10.6 \\
Gemini 3.1 Pro & 22.5 & 36.2 & 22.8 & 8.6 & 23.0 & 38.6 & 23.2 & 7.2 & 14.7 & 24.5 & 10.9 & 8.8 \\
\midrule
Mean & 51.1 & 68.1 & 56.9 & 28.3 & 47.7 & 64.8 & 52.1 & 26.3 & 43.4 & 60.7 & 45.4 & 24.2 \\
\bottomrule\end{tabular}
\end{table}

\label{sec:full_category_scores}
\label{sec:category_supplement}
Figure~\ref{fig:appendixcategories} reports all eleven category scores. For most models, object presence scores exceed spatial-relation scores, subject motion exceeds action events, and camera targeting exceeds camera rotation. Table~\ref{tab:constraint_gaps} summarizes these differences with equal configuration weights. Each category aggregates its applicable checks, while the three consistency dimensions retain equal weight in the overall score.

\begin{table}[htbp]
\centering\small
\setlength{\tabcolsep}{7pt}\renewcommand{\arraystretch}{1.3}
\caption{Differences between category and dimension means, averaged equally across 13 configurations (points). Appearance checks are not applicable to Proxy.}
\label{tab:constraint_gaps}
\begin{tabular}{@{}lrrr@{}}\toprule
Contrast & Proxy & Simplified & Detailed \\ \midrule
Motion $-$ action & 22.2 & 18.3 & 15.7 \\
Target $-$ rotation & 35.0 & 33.1 & 32.1 \\
Presence $-$ relation & 28.3 & 23.4 & 22.4 \\
Presence $-$ appearance & --- & 21.3 & 30.5 \\
Static $-$ dynamic & 28.6 & 25.7 & 21.2 \\
\bottomrule\end{tabular}
\end{table}

\noindent\begin{minipage}{\linewidth}\centering
\includegraphics[width=\linewidth]{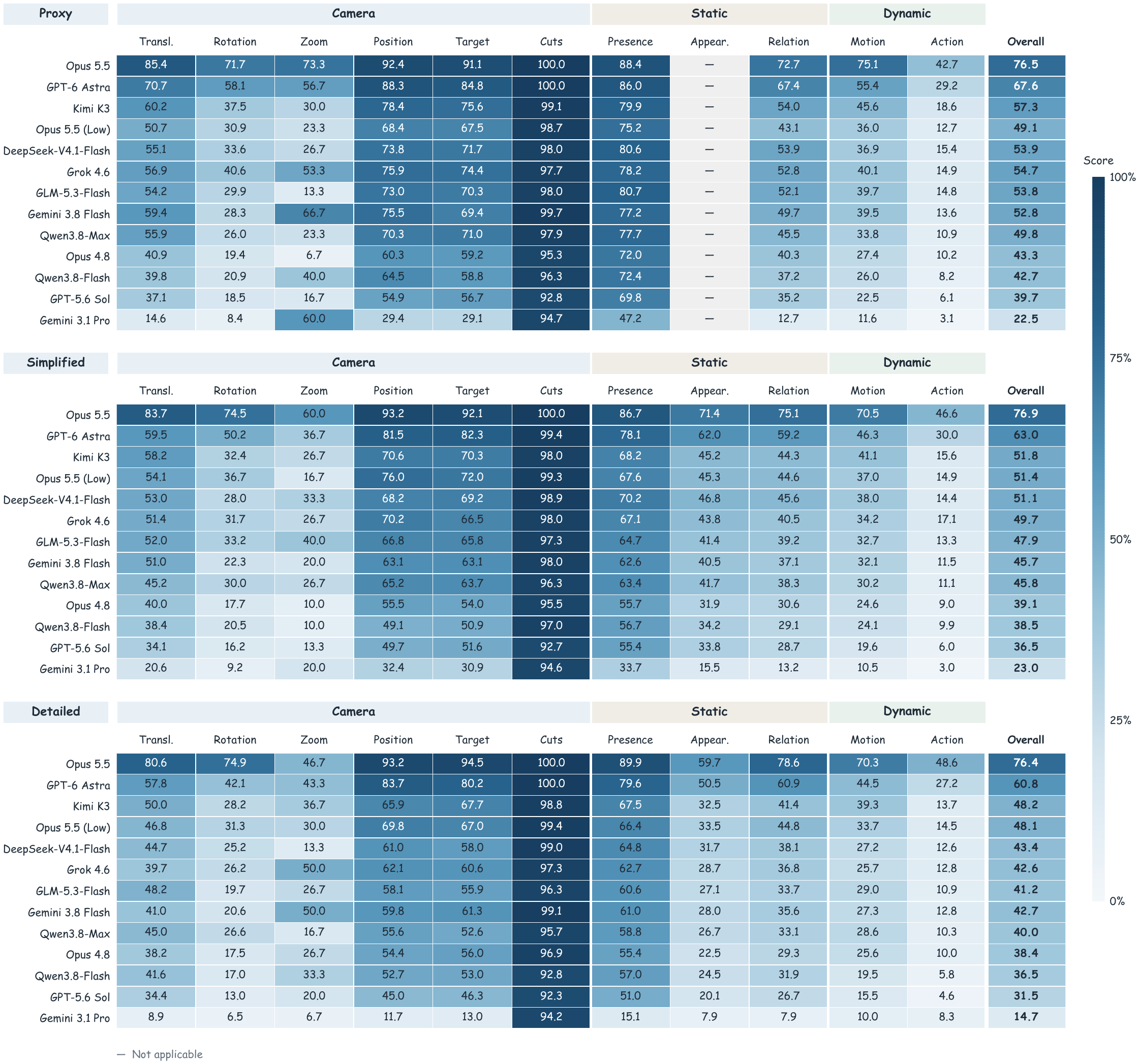}
\captionof{figure}{Complete category scores under the three representation settings. Each cell reports a configuration score on a 0--100 scale, averaging items within clips and then weighting clips equally. Overall is the equal-weight mean of camera, static, and dynamic consistency scores. Panels share configuration order and the 0--100\% color scale shown at right. Low denotes low reasoning effort. Appearance is not applicable to Proxy.}
\label{fig:appendixcategories}
\end{minipage}

\clearpage
\subsection{Additional Reconstructions}
\label{sec:additional_reconstructions}
We compare six models on identical clips and sampled times, followed by selected examples of spatial errors.

\noindent\begin{minipage}{\linewidth}\centering
\captionsetup{font=small,skip=3pt}
\includegraphics[width=.93\linewidth]{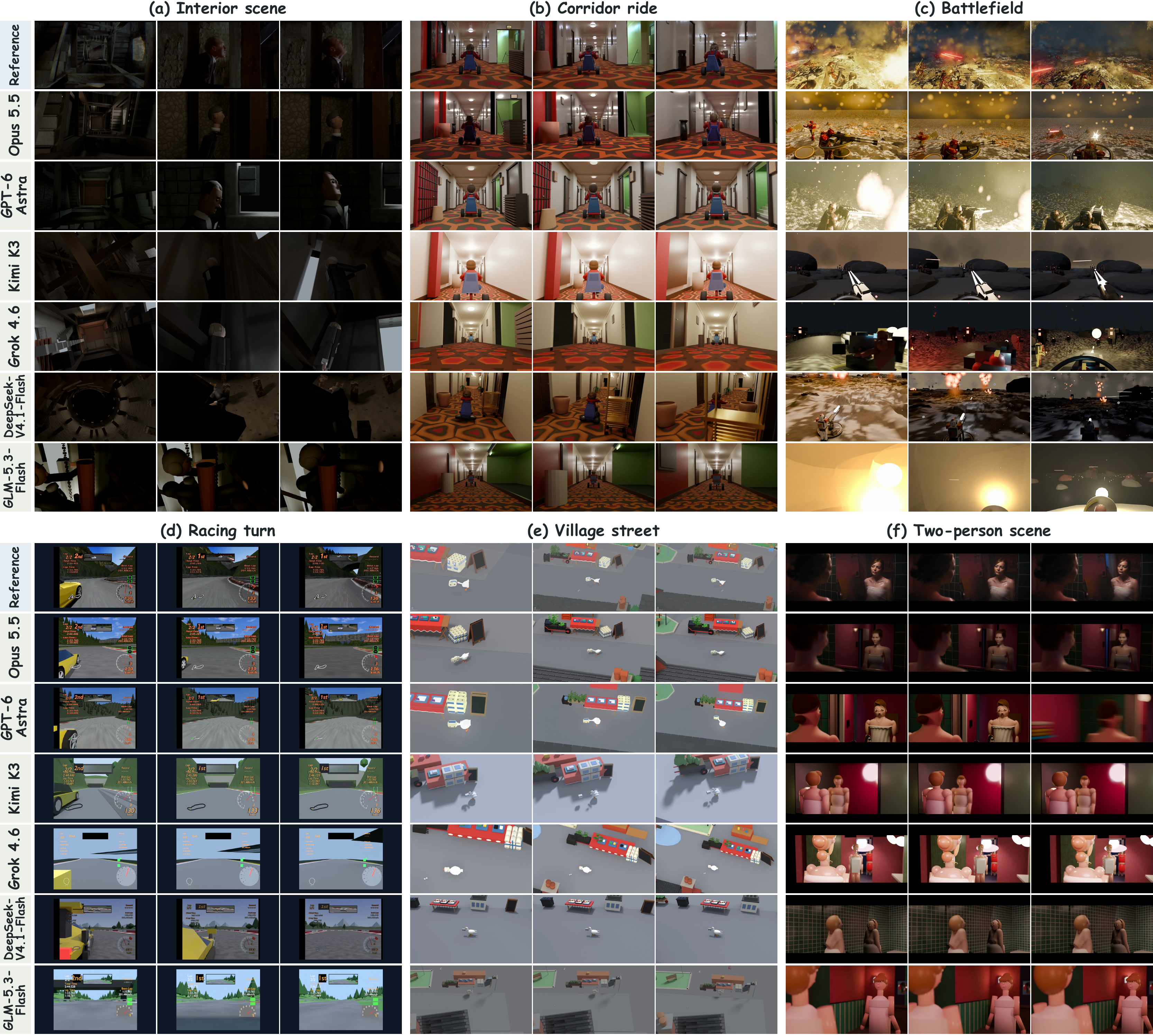}
\captionof{figure}{Detailed reconstructions from six models on six clips. Each column contains three corresponding frames, ordered left to right. Model order is fixed across both blocks.}
\label{fig:appendixrecon1}
\par\vspace{6pt}
\includegraphics[width=.93\linewidth]{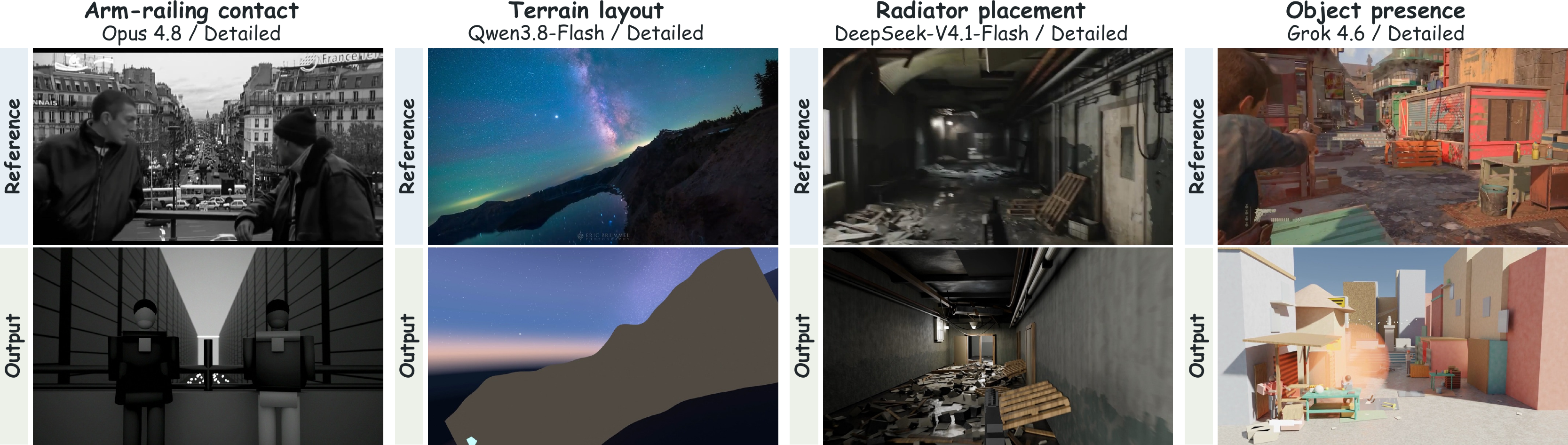}
\captionof{figure}{Selected Detailed errors: arm--railing contact, terrain layout, radiator placement, and object presence. Each column pairs a reference frame with the named model output.}
\label{fig:appendixstatic}
\end{minipage}
\clearpage
\subsection{Comparisons across Representation Settings}
\label{sec:additional_settings}
The same model rows and reference times are retained across Proxy, Simplified, and Detailed for direct comparison.

\noindent\begin{minipage}{\linewidth}\centering
\captionsetup{font=small,skip=3pt}
\includegraphics[width=.93\linewidth]{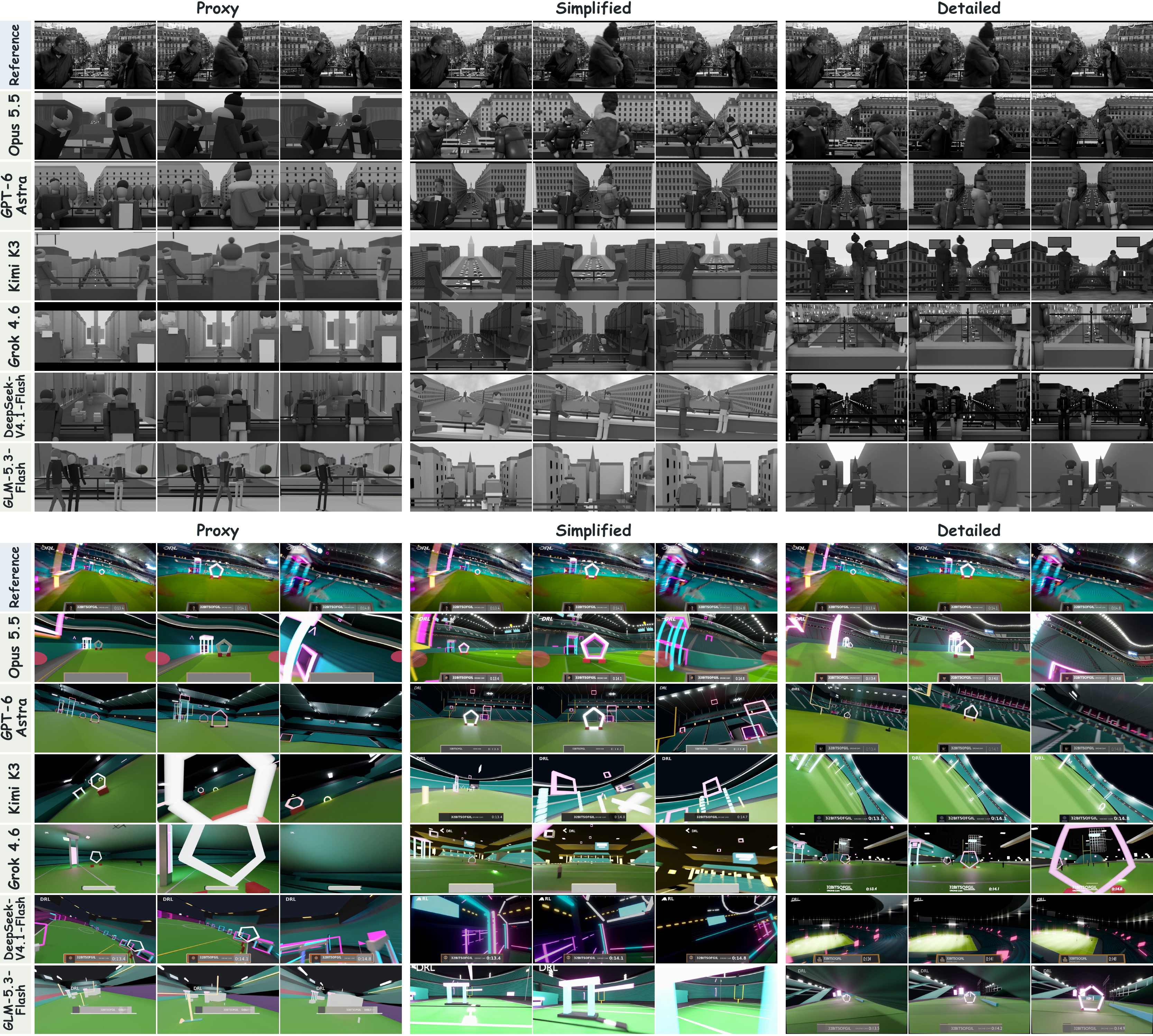}
\captionof{figure}{Six-model comparisons across Proxy, Simplified, and Detailed. The upper and lower blocks show a street conversation and drone racing, respectively. All three settings share the reference times within each block.}
\label{fig:appendixsettings1}
\par\vspace{6pt}
\includegraphics[width=.93\linewidth]{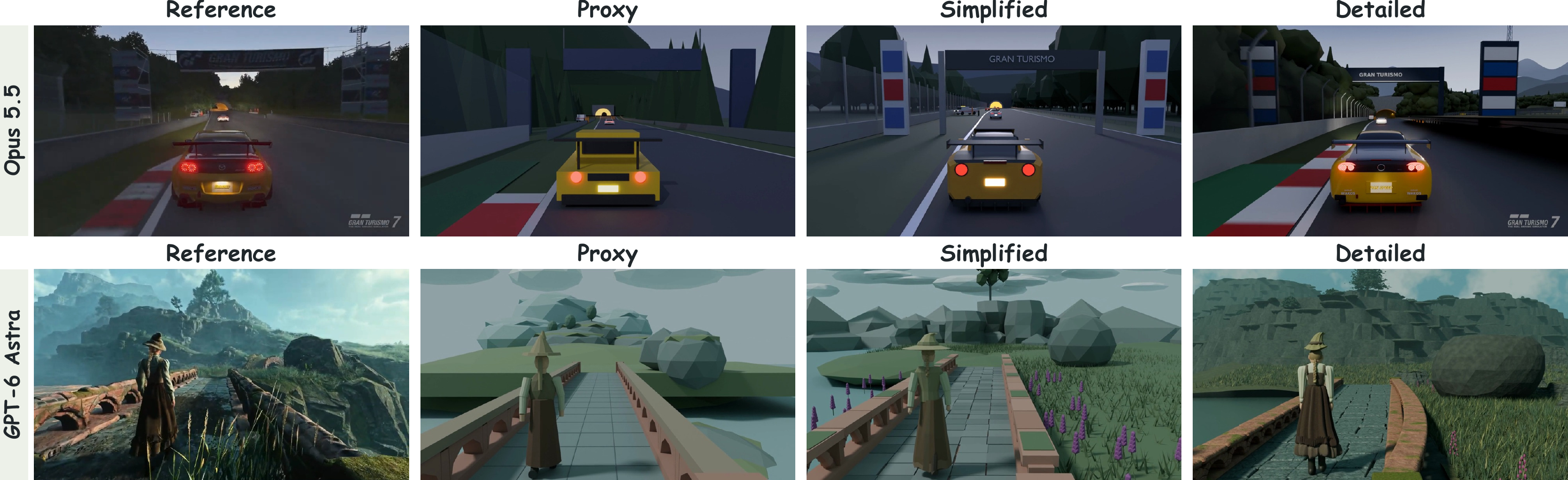}
\captionof{figure}{Selected setting comparisons at enlarged scale: vehicle racing with Opus 5.5 (top) and a mountain landscape with GPT-6 Astra (bottom). Each row fixes the model and reference time.}
\label{fig:appendixsettings2}
\end{minipage}
\clearpage
\subsection{Actions and Camera Changes}
\label{sec:additional_failures}
The Proxy comparison exposes action progression across six models. Selected Detailed examples complement it with camera turns and shot changes.

\noindent\begin{minipage}{\linewidth}\centering
\captionsetup{font=small,skip=3pt}
\includegraphics[width=.93\linewidth]{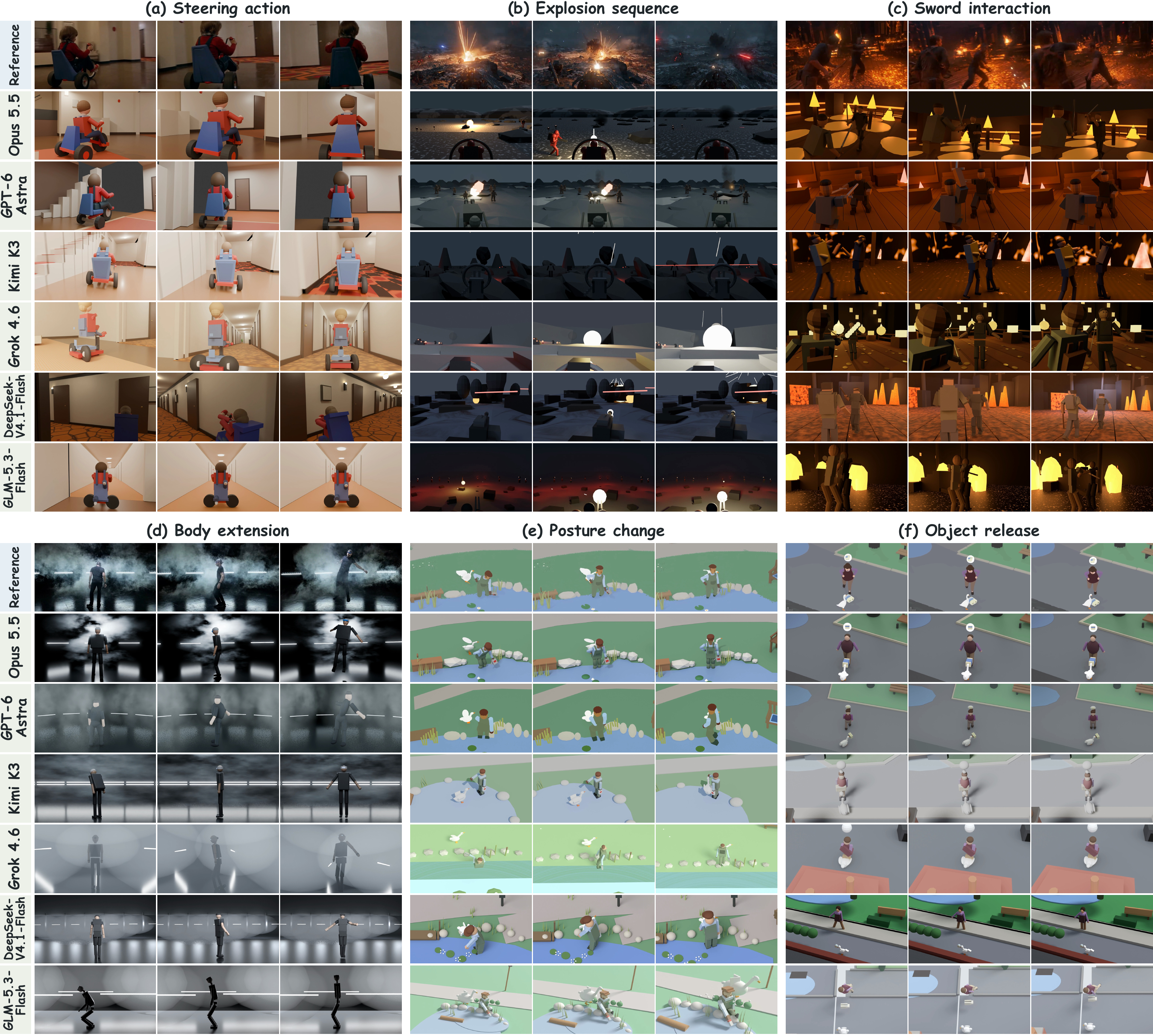}
\captionof{figure}{Proxy action sequences from six models: steering, explosions, sword interaction, body extension, posture change, and object release. Three corresponding frames progress left to right in each column.}
\label{fig:appendixmotion}
\par\vspace{6pt}
\includegraphics[width=.93\linewidth]{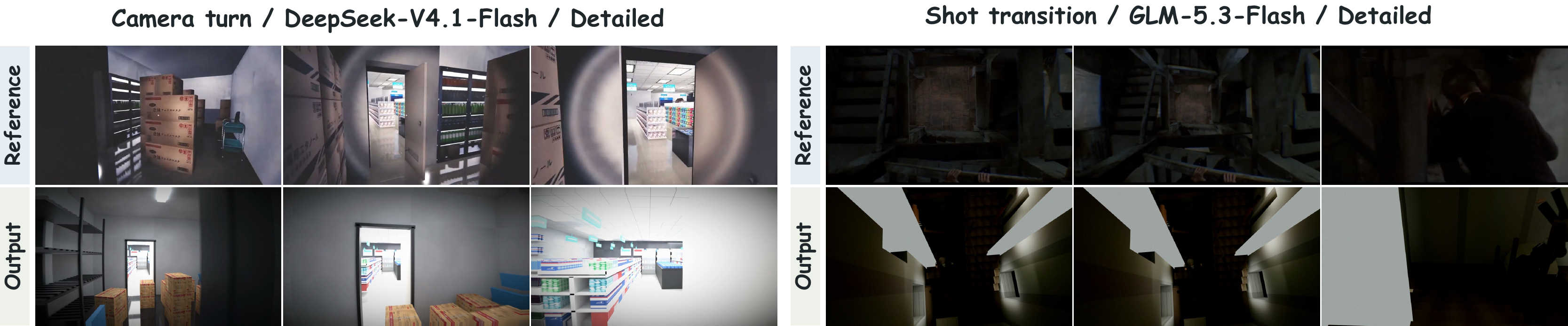}
\captionof{figure}{Selected Detailed camera errors: an indoor turn (DeepSeek-V4.1-Flash, left) and a shot transition (GLM-5.3-Flash, right). Reference and output share the same three sampled times.}
\label{fig:appendixcamera}
\end{minipage}
\clearpage
\end{document}

%% file: preamble.tex
\usepackage{iftex}
\ifXeTeX
  \usepackage{fontspec}
\fi

\usepackage{cuted} 

\usepackage{currfile} 

\usepackage{caption} 

\usepackage{tcolorbox}
\tcbuselibrary{breakable,skins}
\newtcolorbox{promptbox}[1]{
  enhanced,breakable,sharp corners,
  colback=black!2,colframe=black!25,colbacktitle=black!7,coltitle=black,
  boxrule=0.4pt,titlerule=0.3pt,
  left=8pt,right=8pt,top=6pt,bottom=6pt,
  toptitle=4pt,bottomtitle=4pt,
  before skip=8pt,after skip=8pt,
  fonttitle=\small\sffamily\bfseries,
  fontupper=\fontsize{9}{10}\selectfont\ttfamily,
  before upper={\raggedright\setlength{\parskip}{5pt}},
  title={#1}}

%% file: fig/teaser.tex
\begin{strip}
\vspace{-54pt}
\centering
\includegraphics[width=\linewidth]{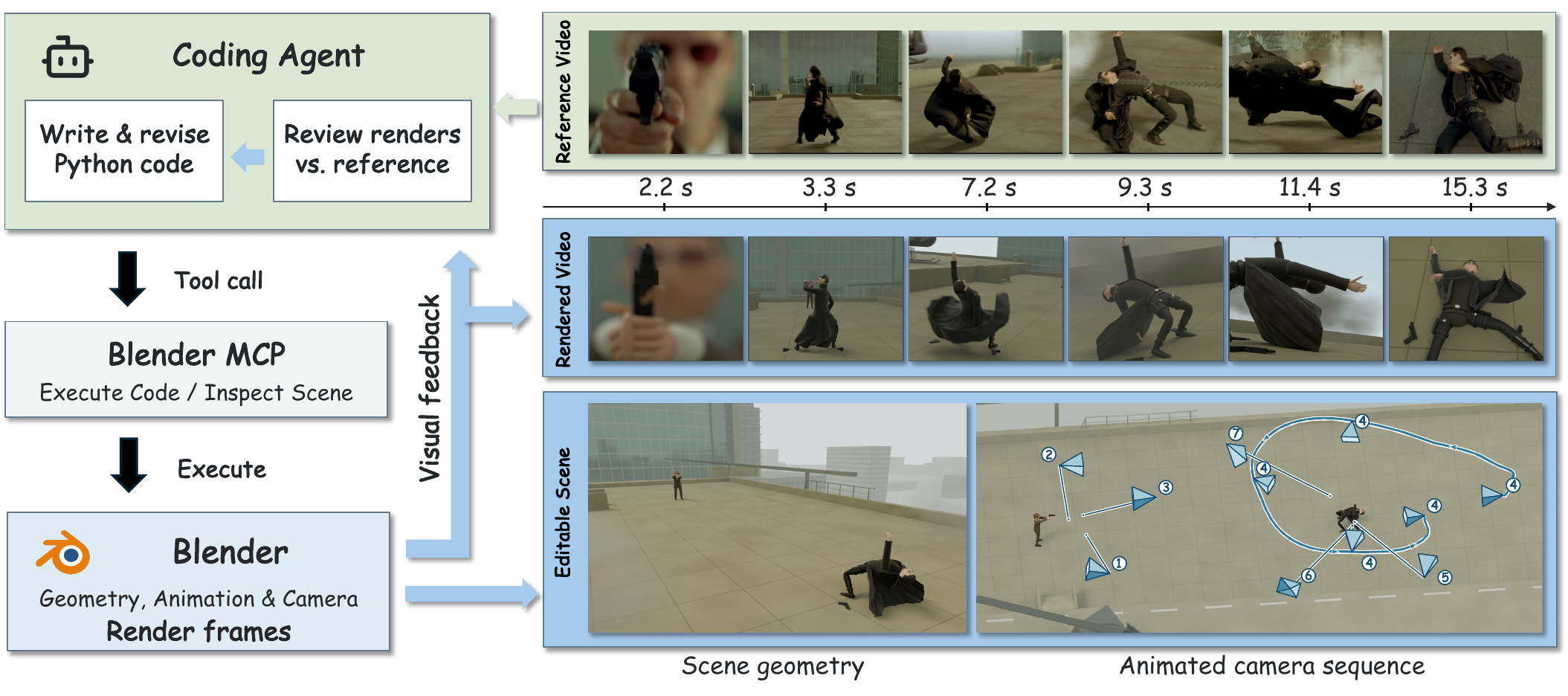}
\captionsetup{hypcap=false}
\captionof{figure}{\textbf{CineRecon task overview.} Given a complete reference video, an agent writes Python code and executes it through Blender MCP to reconstruct an editable dynamic scene, compares rendered frames with the reference, and iteratively revises the code. The Opus 5.5 Detailed example shows aligned reference and rendered frames, scene geometry, and the animated camera sequence. Camera labels indicate shot order.}
\captionsetup{hypcap=true}
\label{fig:benchmark_overview}
\end{strip}

%% file: references.bib
@inproceedings{gu2025blendergym,
  title={Blendergym: Benchmarking foundational model systems for graphics editing},
  author={Gu, Yunqi and Huang, Ian and Je, Jihyeon and Yang, Guandao and Guibas, Leonidas},
  booktitle={2025 IEEE/CVF Conference on Computer Vision and Pattern Recognition (CVPR)},
  pages={18574--18583},
  year={2025},
  organization={IEEE}
}

@inproceedings{yin2026vision,
  title={Vision-as-inverse-graphics agent via interleaved multimodal reasoning},
  author={Yin, Shaofeng and Ge, Jiaxin and Wang, Zora Zhiruo and Wang, Chenyang and Li, Xiuyu and Black, Michael J and Darrell, Trevor and Kanazawa, Angjoo and Feng, Haiwen},
  booktitle={European Conference on Computer Vision},
  pages={557--575},
  year={2026},
  organization={Springer}
}

@article{liu20263dharnessbench,
  title={3DHarnessBench: Probing Agentic 3D-to-Code Capabilities of Frontier Vision-Language Models},
  author={Liu, Ling and Gong, Bingchen and Parakkat, Amal Dev and Ovsjanikov, Maks},
  journal={arXiv preprint arXiv:2609.06535},
  year={2026}
}

@article{tang2026bvb,
  title={BVB: Benchmarking Agentic Video Understanding via Programmatic Reconstruction in Blender},
  author={Tang, Yolo Y and Shimada, Daiki and Meng, Jiayue and Bi, Jing and Liu, Pinxin and Wang, Yicheng and Xiao, Yunzhong and Tan, Zhangyun and Zhang, Zeliang and Huang, Chao and others},
  journal={arXiv preprint arXiv:2609.15478},
  year={2026}
}

@article{zhao2026sceneactbench,
  title={SceneActBench: Can Agents Act on the 3D Scenes They See?},
  author={Zhao, Yifei and Zhou, Xiangxin and Yang, Wenhao and Tang, Jiaqi and Jian, Pu and Yao, Huanjin and Yao, Jiarui and Lin, Haowei and Guo, Chunchao and Chen, Zhuo and others},
  journal={arXiv preprint arXiv:2607.22393},
  year={2026}
}

@article{liang2026visphyworld,
  title={VisPhyWorld: Probing Physical Reasoning via Code-Driven Video Reconstruction},
  author={Liang, Jiarong and Ku, Max and Hui, Ka-Hei and Nie, Ping and Chen, Wenhu},
  journal={arXiv preprint arXiv:2602.13294},
  year={2026}
}

@inproceedings{huang2024vbench,
  title={Vbench: Comprehensive benchmark suite for video generative models},
  author={Huang, Ziqi and He, Yinan and Yu, Jiashuo and Zhang, Fan and Si, Chenyang and Jiang, Yuming and Zhang, Yuanhan and Wu, Tianxing and Jin, Qingyang and Chanpaisit, Nattapol and others},
  booktitle={2024 IEEE/CVF Conference on Computer Vision and Pattern Recognition (CVPR)},
  pages={21807--21818},
  year={2024},
  organization={IEEE}
}

@article{lin2026towards,
  title={Towards understanding camera motions in any video},
  author={Lin, Zhiqiu and Cen, Siyuan and Jiang, Daniel and Karhade, Jay and Wang, Hewei and Mitra, Chancharik and Ling, Yu Tong Tiffany and Huang, Yuhan and Zawar, Rushikesh and Bai, Xue and others},
  journal={Advances in Neural Information Processing Systems},
  volume={38},
  year={2026}
}

@article{shen20264dcodebench,
  title={4DCodeBench: Benchmarking Agents on Inverse Graphics of Dynamic Scenes},
  author={Shen, Ruihong and Kova{\v{c}}i{\v{c}}, {\v{Z}}iga and Kulits, Peter and Wang, Xingrui and Li, Zizhang and Tenenbaum, Joshua B and Yuille, Alan and Chen, Jieneng and Wu, Jiajun},
  journal={arXiv preprint arXiv:2610.03715},
  year={2026}
}

@article{gao20263dcodebench,
  title={3DCodeBench: Benchmarking Agentic Procedural 3D Modeling Via Code},
  author={Gao, Yipeng and Shu, Lei and Ye, Genzhi and Xiong, Xi and Makadia, Ameesh and Guo, Meiqi and Itti, Laurent and Chen, Jindong},
  journal={arXiv preprint arXiv:2606.01057},
  year={2026}
}

@article{liu2026ir3d,
  title={Ir3d-bench: Evaluating vision-language model scene understanding as agentic inverse rendering},
  author={Liu, Hengyu and Li, Chenxin and Li, Zhengxin and Wu, Yipeng and Li, Wuyang and Yang, Zhiqin and Zhang, Zhenyuan and Lin, Yunlong and Han, Sirui and Feng, Brandon},
  journal={Advances in Neural Information Processing Systems},
  volume={38},
  year={2026}
}

@inproceedings{liu2025shape,
  title={Shape-for-motion: Precise and consistent video editing with 3d proxy},
  author={Liu, Yuhao and Wang, Tengfei and Liu, Fang and Wang, Zhenwei and Lau, Rynson WH},
  booktitle={Proceedings of the SIGGRAPH Asia 2025 Conference Papers},
  pages={1--12},
  year={2025}
}

@article{fang2026spatialcrafter,
  title={SpatialCrafter: Single Image World Modeling with Generative 3D Proxies},
  author={Fang, Chuan and Qiu, Lingteng and Liang, Yixun and Chen, Rui and Luo, Kunming and Zheng, Zhaohua and Bai, Tongyuan and Tian, Feipeng and Dong, Zilong and Zhou, Zihan and others},
  journal={arXiv preprint arXiv:2608.27073},
  year={2026}
}

@inproceedings{ling2025vmbench,
  title={Vmbench: A benchmark for perception-aligned video motion generation},
  author={Ling, Xinran and Zhu, Chen and Wu, Meiqi and Li, Hangyu and Feng, Xiaokun and Yang, Cundian and Hao, Aiming and Zhu, Jiashu and Wu, Jiahong and Chu, Xiangxiang},
  booktitle={2025 IEEE/CVF International Conference on Computer Vision (ICCV)},
  pages={13087--13098},
  year={2025},
  organization={IEEE}
}

@inproceedings{wang2024motionctrl,
  title={Motionctrl: A unified and flexible motion controller for video generation},
  author={Wang, Zhouxia and Yuan, Ziyang and Wang, Xintao and Li, Yaowei and Chen, Tianshui and Xia, Menghan and Luo, Ping and Shan, Ying},
  booktitle={ACM SIGGRAPH 2024 Conference Papers},
  pages={1--11},
  year={2024}
}

@article{he2024cameractrl,
  title={Cameractrl: Enabling camera control for text-to-video generation},
  author={He, Hao and Xu, Yinghao and Guo, Yuwei and Wetzstein, Gordon and Dai, Bo and Li, Hongsheng and Yang, Ceyuan},
  journal={arXiv preprint arXiv:2404.02101},
  year={2024}
}

@inproceedings{huang2024blenderalchemy,
  title={Blenderalchemy: Editing 3d graphics with vision-language models},
  author={Huang, Ian and Yang, Guandao and Guibas, Leonidas},
  booktitle={European Conference on Computer Vision},
  pages={297--314},
  year={2024},
  organization={Springer}
}

@article{feng2024tc,
  title={Tc-bench: Benchmarking temporal compositionality in text-to-video and image-to-video generation},
  author={Feng, Weixi and Li, Jiachen and Saxon, Michael and Fu, Tsu-jui and Chen, Wenhu and Wang, William Yang},
  journal={arXiv preprint arXiv:2406.08656},
  year={2024}
}

@inproceedings{hu2024scenecraft,
  title={SceneCraft: An LLM Agent for Synthesizing 3D Scenes as Blender Code.},
  author={Hu, Ziniu and Iscen, Ahmet and Jain, Aashi and Kipf, Thomas and Yue, Yisong and Ross, David A and Schmid, Cordelia and Fathi, Alireza},
  booktitle={ICML},
  pages={19252--19282},
  year={2024}
}

@article{kulits2024re,
  title={Re-thinking inverse graphics with large language models},
  author={Kulits, Peter and Feng, Haiwen and Liu, Weiyang and Abrevaya, Victoria and Black, Michael J},
  journal={arXiv preprint arXiv:2404.15228},
  year={2024}
}

@article{he2026thinking,
  title={Thinking in Blender: Staged Executable Inverse Graphics with Vision-Language Models},
  author={He, Guangzhao and Luo, Rundong and Ma, Wei-Chiu and Averbuch-Elor, Hadar},
  journal={arXiv preprint arXiv:2606.02580},
  year={2026}
}

@article{zhang2026code2worlds,
  title={Code2worlds: Empowering coding llms for 4d world generation},
  author={Zhang, Yi and Wang, Yunshuang and Zhang, Zeyu and Tang, Hao},
  journal={arXiv preprint arXiv:2602.11757},
  year={2026}
}

@article{liu2026simworlds,
  title={SimWorlds: A Multi-Agent System for Dynamic 3D Scene Creation},
  author={Liu, Chunjiang and Wang, Xiaoyuan and Chen, Haoyu and Zhao, Yizhou and Yang, Ming-Hsuan and Jeni, L{\'a}szl{\'o} A},
  journal={arXiv preprint arXiv:2607.01766},
  year={2026}
}

@inproceedings{liu2024physgen,
  title={Physgen: Rigid-body physics-grounded image-to-video generation},
  author={Liu, Shaowei and Ren, Zhongzheng and Gupta, Saurabh and Wang, Shenlong},
  booktitle={European Conference on Computer Vision},
  pages={360--378},
  year={2024},
  organization={Springer}
}

@inproceedings{hu2023tifa,
  title={Tifa: Accurate and interpretable text-to-image faithfulness evaluation with question answering},
  author={Hu, Yushi and Liu, Benlin and Kasai, Jungo and Wang, Yizhong and Ostendorf, Mari and Krishna, Ranjay and Smith, Noah A},
  booktitle={2023 ieee/cvf international conference on computer vision (iccv)},
  pages={20349--20360},
  year={2023},
  organization={IEEE}
}

@inproceedings{cho2024davidsonian,
  title={Davidsonian scene graph: Improving reliability in fine-grained evaluation for text-to-image generation},
  author={Cho, Jaemin and Hu, Yushi and Baldridge, Jason and Garg, Roopal and Anderson, Peter and Krishna, Ranjay and Bansal, Mohit and Pont-Tuset, Jordi and Wang, Su},
  booktitle={International conference on learning representations},
  volume={2024},
  pages={15625--15645},
  year={2024}
}

@inproceedings{ku2024viescore,
  title={Viescore: Towards explainable metrics for conditional image synthesis evaluation},
  author={Ku, Max and Jiang, Dongfu and Wei, Cong and Yue, Xiang and Chen, Wenhu},
  booktitle={Proceedings of the 62nd Annual Meeting of the Association for Computational Linguistics (Volume 1: Long Papers)},
  pages={12268--12290},
  year={2024}
}

@inproceedings{he2024videoscore,
  title={Videoscore: Building automatic metrics to simulate fine-grained human feedback for video generation},
  author={He, Xuan and Jiang, Dongfu and Zhang, Ge and Ku, Max and Soni, Achint and Siu, Sherman and Chen, Haonan and Chandra, Abhranil and Jiang, Ziyan and Arulraj, Aaran and others},
  booktitle={Proceedings of the 2024 Conference on Empirical Methods in Natural Language Processing},
  pages={2105--2123},
  year={2024}
}

@inproceedings{bansal2024videocon,
  title={Videocon: Robust video-language alignment via contrast captions},
  author={Bansal, Hritik and Bitton, Yonatan and Szpektor, Idan and Chang, Kai-Wei and Grover, Aditya},
  booktitle={2024 IEEE/CVF Conference on Computer Vision and Pattern Recognition (CVPR)},
  pages={13927--13937},
  year={2024},
  organization={IEEE}
}

@misc{blender2026,
  author       = {{Blender Foundation}},
  title        = {{Blender}},
  year         = {2026},
  url          = {https://www.blender.org/},
  howpublished = {Computer software}
}

@misc{ahuja2025blendermcp,
  author       = {Ahuja, Siddharth},
  title        = {{MCP for Blender (formerly Blender MCP)}},
  year         = {2025},
  url          = {https://github.com/ahujasid/mcp-for-blender},
  howpublished = {Computer software}
}

@misc{openai2026astra,
  author = {{OpenAI}},
  title = {{GPT-6 Astra}},
  year = {2026},
  howpublished = {\url{https://developers.openai.com/api/docs/models/gpt-6-astra}},
  note = {Accessed October 8, 2026},
  url = {https://developers.openai.com/api/docs/models/gpt-6-astra}
}

@misc{openai2026sol,
  author = {{OpenAI}},
  title = {{GPT-5.6 Sol}},
  year = {2026},
  howpublished = {\url{https://developers.openai.com/api/docs/models/gpt-5.6-sol}},
  note = {Accessed October 8, 2026},
  url = {https://developers.openai.com/api/docs/models/gpt-5.6-sol}
}

@misc{anthropic2026opus55,
  author = {{Anthropic}},
  title = {{Introducing Claude Opus 5.5}},
  year = {2026},
  howpublished = {\url{https://www.anthropic.com/claude-opus-5-5}},
  note = {Accessed October 8, 2026},
  url = {https://www.anthropic.com/claude-opus-5-5}
}

@misc{anthropic2026opus48,
  author = {{Anthropic}},
  title = {{Introducing Claude Opus 4.8}},
  year = {2026},
  howpublished = {\url{https://www.anthropic.com/news/claude-opus-4-8}},
  note = {Accessed October 8, 2026},
  url = {https://www.anthropic.com/news/claude-opus-4-8}
}

@misc{google2026gemini31,
  author = {{Google DeepMind}},
  title = {{Gemini 3.1 Pro: Model Card}},
  year = {2026},
  howpublished = {\url{https://deepmind.google/models/model-cards/gemini-3-1-pro/}},
  note = {Accessed October 8, 2026},
  url = {https://deepmind.google/models/model-cards/gemini-3-1-pro/}
}

@misc{google2026gemini38,
  author = {{Google DeepMind}},
  title = {{Gemini 3.8 Flash: Model Card}},
  year = {2026},
  howpublished = {\url{https://deepmind.google/models/model-cards/gemini-3-8-flash/}},
  note = {Accessed October 8, 2026},
  url = {https://deepmind.google/models/model-cards/gemini-3-8-flash/}
}

@misc{alibaba2026qwenmax,
  author = {{Alibaba Cloud}},
  title = {{Qwen3.8-Max: Model Information}},
  year = {2026},
  howpublished = {\url{https://www.alibabacloud.com/help/en/model-studio/qwen3-8-max}},
  note = {Accessed October 8, 2026},
  url = {https://www.alibabacloud.com/help/en/model-studio/qwen3-8-max}
}

@misc{alibaba2026qwenflash,
  author = {{Alibaba Cloud}},
  title = {{Qwen3.8-Flash: Model Information}},
  year = {2026},
  howpublished = {\url{https://www.alibabacloud.com/help/en/model-studio/qwen3-8-flash}},
  note = {Accessed October 8, 2026},
  url = {https://www.alibabacloud.com/help/en/model-studio/qwen3-8-flash}
}

@misc{moonshot2026kimik3,
  author = {{Moonshot AI}},
  title = {{Kimi-K3: Model Card}},
  year = {2026},
  howpublished = {\url{https://huggingface.co/moonshotai/Kimi-K3}},
  note = {Accessed October 8, 2026},
  url = {https://huggingface.co/moonshotai/Kimi-K3}
}

@misc{xai2026grok46,
  author = {{SpaceXAI}},
  title = {{Introducing Grok 4.6}},
  year = {2026},
  howpublished = {\url{https://x.ai/news/grok-4-6}},
  note = {Accessed October 8, 2026},
  url = {https://x.ai/news/grok-4-6}
}

@misc{deepseek2026v41,
  author = {{DeepSeek}},
  title = {{DeepSeek-V4.1-Flash: Smarter, Faster, More Efficient}},
  year = {2026},
  howpublished = {\url{https://api-docs.deepseek.com/news/news260910/}},
  note = {Accessed October 8, 2026},
  url = {https://api-docs.deepseek.com/news/news260910/}
}

@misc{zai2026glm53,
  author = {{Z.ai}},
  title = {{GLM-5.3-Flash: Model Card}},
  year = {2026},
  howpublished = {\url{https://huggingface.co/zai-org/GLM-5.3-Flash}},
  note = {Accessed October 8, 2026},
  url = {https://huggingface.co/zai-org/GLM-5.3-Flash}
}
